%% file: main.tex
\documentclass[letterpaper, 10 pt, journal, twoside]{IEEEtran}
\ifCLASSINFOpdf
\else
\fi
\usepackage{amsmath,amsfonts}
\usepackage{array}
\usepackage{textcomp}
\usepackage{stfloats}
\usepackage{url}
\usepackage{verbatim}
\usepackage{graphicx}
\usepackage{cite}
\usepackage{graphics} 
\usepackage{epsfig} 
\usepackage{empheq}
\usepackage{times} 
\usepackage{amssymb}  
\usepackage{bm}
\usepackage{color}
\usepackage{soul}
\usepackage{bbding} 
\usepackage{diagbox}
\usepackage[linesnumbered,ruled]{algorithm2e}
\usepackage{ulem} 
\usepackage{hyperref}
\hypersetup{
colorlinks=true,
linkcolor=blue,
filecolor=blue,      
urlcolor=blue,
citecolor=cyan,
}

\usepackage{float}
\usepackage{booktabs}
\usepackage{multirow}
\usepackage{mathrsfs}
\usepackage[utf8]{inputenc}
\usepackage{pifont}
\usepackage{threeparttable}
\usepackage{caption}
\usepackage{setspace}
\newcounter{RNum}

\input{packages}
\begin{document}

\author{Chao Chen$^1$, Zegang Cheng$^{1,*}$, Xinhao Liu$^{1,*}$, Yiming Li$^1$, Li Ding$^2$, Ruoyu Wang$^1$, and Chen Feng\textsuperscript{1,\ding{41}}
\\
\thanks{Manuscript received: Month Day, Year; Revised: June 24, 2024; Accepted: November 1, 2024. This paper was recommended for publication by Editor Cesar Cadena Lerma upon evaluation of the Associate Editor and Reviewers’ comments.}
\thanks{The work was supported in part through NSF 2238968, 2322242, 2121391, and 2036870 grants. We thank Xuchu Xu for insightful discussions.}
\thanks{* Equal contribution}
\thanks{\ding{41} Corresponding author.}
\thanks{$^{1}$Chao Chen, Zegang Cheng, Xinhao Liu, Yiming Li, Ruoyu Wang, and Chen Feng are with New York University,
Brooklyn, NY 11201, USA  {\tt\small cfeng@nyu.edu}}
\thanks{$^{2}$Li Ding is with University of Rochester, Rochester, NY 14627, USA {\tt\small l.ding@rochester.edu}}

\thanks{The webpage of this paper is available at https://ai4ce.github.io/TF-VPR/.}
\thanks{Digital Object Identifier(DOI): see top of this page.}
}

\title{\LARGE \bf Self-Supervised Place Recognition by Refining \\ Temporal and Featural Pseudo Labels from Panoramic Data} 

\markboth{IEEE Robotics and Automation Letters. Preprint Version. Accepted Nov, 2024}
{Chen \MakeLowercase{\textit{et al.}}: Self-Supervised Place Recognition by Refining Temporal and Featural Pseudo Labels from Panoramic Data}

\maketitle
\bstctlcite{IEEEexample:BSTcontrol}
\input{parts/0-title-author-abstract}

\input{parts/1-intro}
\input{parts/2-related}
\input{parts/3-method}
\input{parts/4-experiments}
\input{parts/5-conclusions}

\bibliography{IEEEabrv,egbib}
\clearpage
\input{parts/Appendix}

\bibliographystyle{IEEEtran}  
\normalem

\end{document}

%% file: packages.tex
\usepackage{xcolor}
\usepackage{subcaption}
\usepackage{comment}
\usepackage{wrapfig}

\usepackage{dsfont}



%% file: parts/0-title-author-abstract.tex
\providecommand{\titlevariable}{TF-VPR}
\providecommand{\realworld}{KITTI-360}
\begin{abstract}
Visual place recognition (VPR) using deep networks has achieved state-of-the-art performance. However, most of them require a training set with ground truth sensor poses to obtain positive and negative samples of each observation’s spatial neighborhood for supervised learning. When such information is unavailable, temporal neighborhoods from a sequentially collected data stream could be exploited for self-supervised training, although we find its performance suboptimal. Inspired by noisy label learning, we propose a novel self-supervised framework named TF-VPR that uses temporal neighborhoods and learnable feature neighborhoods to discover unknown spatial neighborhoods. Our method follows an iterative training paradigm which alternates between: (1) representation learning with data augmentation, (2) positive set expansion to include the current feature space neighbors, and (3) positive set contraction via geometric verification. We conduct auto-labeling and generalization tests on both simulated and real datasets, with either RGB images or point clouds as inputs. The results show that our method outperforms self-supervised baselines in recall rate, robustness, and heading diversity, a novel metric we propose for VPR. Our code and datasets can be found at \url{https://ai4ce.github.io/TF-VPR/}.

\end{abstract}

%% file: parts/1-intro.tex
\section{Introduction}



Visual place recognition (VPR), which aims to identify previously visited places based on current visual observation, is a well-known problem in computer vision and plays a crucial role in autonomous robots. Meanwhile, VPR is closely related to re-localization, loop closure detection, and image retrieval. Despite all efforts, VPR remains a difficult task due to various challenges such as perceptual aliasing and view direction differences~\cite{zaffar2021vpr,sheng2021nyu}. Classic VPR methods based on hand-crafted feature matching do not require supervised learning, but are less robust to the challenges mentioned above~\cite{lowe1999object,arandjelovic2013all}. Thus, learning-based methods have been proposed to learn local or global feature descriptors~\cite{sarlin2019coarse} by place classification~\cite{chen2017deep} or contrastive-like similarity learning~\cite{arandjelovic2016netvlad}. Some works also use a sequence of images instead of a single image to mitigate perceptual aliasing issues.~\cite{garg2022seqmatchnet,vysotska2015lazy}. 

So far, most learning-based VPR methods are supervised, focusing on either learning better feature representations or designing robust matching strategies. They assume that ground truth camera poses are available in their training sets, for obtaining the positive and negative training triplets of each visual observation~\cite{arandjelovic2016netvlad,sarlin2019coarse}, or defining place categories~\cite{chen2017deep}. Although noisy ground truth, such as inaccurate GPS signal, are available in outdoor environments, obtaining such labels is rather non-trivial for indoor scenes. One possible approach is to use existing visual SLAM or SfM methods to estimate camera poses. However, these methods can not always guarantee accurate pose estimations due to various challenges including the need for robust loop closing from VPR itself.
Considering human's extraordinary VPR ability that does not seem to need ground truth pose information for its training, we ask the following question: \textit{is it possible to relax such an assumption and design a learning-based VPR approach without pose-dependent supervision?}



\begin{figure}[t]
     \centering
     \begin{subfigure}[t]{0.30\textwidth}
         \centering
         \includegraphics[width=\textwidth]{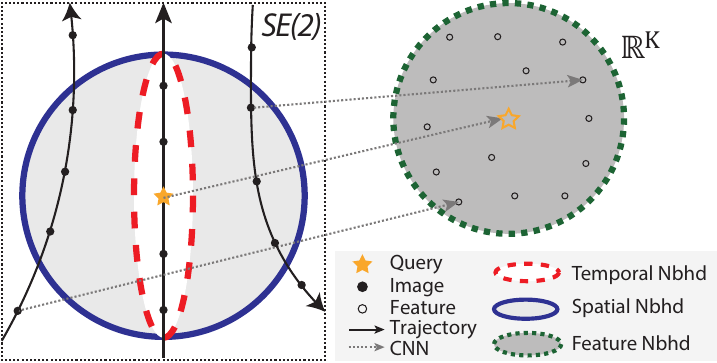}
         \caption{A partial sensory stream \& 3 types of neighbors (\textcolor{red}{red}/\textcolor{blue}{blue}/\textcolor{teal}{green} circles).}
         \label{fig:neighbor1}
     \end{subfigure}
     \begin{subfigure}[t]{0.17\textwidth}
        \centering
        \includegraphics[width=\textwidth]{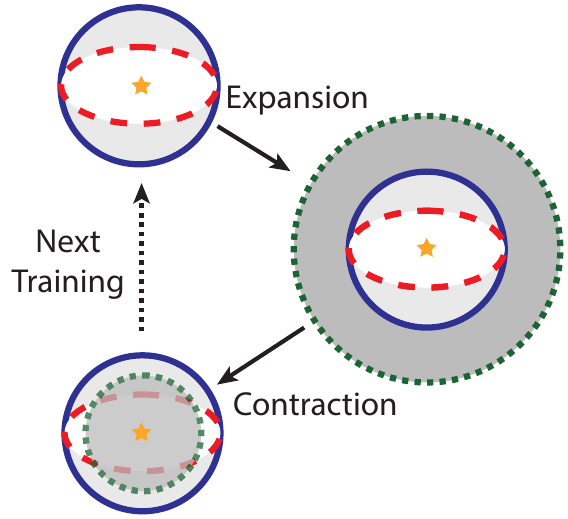}
        \caption{A spatial neighborhood's iterative update.}
        \label{fig:neighbor2}
    \end{subfigure}
     \caption{We introduce the first iterative approach to explore a query's spatial neighbors given its temporal neighbors. Our solution is based on the interconnections between the \textcolor{red}{temporal}, \textcolor{blue}{spatial}, and \textcolor{teal}{feature} neighborhoods in sensory data: a query’s spatial neighborhood expands from its temporal to feature neighbors, then contracts to exclude wrong neighbors, iterated in training until such neighborhoods’ convergence. 
     }
     \vspace{-6mm}
     \label{fig:neighborhoods}
\end{figure}

To achieve this goal, our main idea is \textit{to leverage fixed temporal neighborhoods and learnable feature neighborhoods to discover the unknown spatial neighborhoods} (which require ground truth poses to compute), leading to a self-supervised VPR method shown in Fig.~\ref{fig:neighborhoods}.
We are inspired by research work utilizing sensory streams (RGB videos or point cloud sequences) to obtain the positive and negative neighbors in the temporal domain such as~\cite{savinov2018semi}. However, we find that VPR learned from temporal cues alone will miss spatial neighboring places with large viewpoint differences, because temporal neighbors tend to share similar viewpoints. 

This is suboptimal in applications such as visual navigation or loop closure for SLAM. To automatically discover the true spatial neighbors with diverse viewpoints, we propose a novel iterative learning strategy inspired by noisy label training such as bootstrapping~\cite{reed2014training}. More specifically, we exploit the temporal information for label initialization as shown in Fig.~\ref{fig:neighbor1}. 
Afterward, a feature representation will be learned based on the selected triplets. Then, as shown in Fig.~\ref{fig:neighbor2}, we select the training triplets using the feature space by: (1) adding feature-space neighbors as tentative positives, (2) rejecting false positives and false negatives via geometric verification, in order to further refine the feature representation. Note that the above steps are iteratively conducted until convergence. 

To evaluate our self-supervised VPR with \textbf{T}emporal and \textbf{F}eature neighborhood interactions~(\textbf{\titlevariable)}, we generate a simulated RGB dataset~\cite{savva2019habitat}, and apply a real-world point cloud dataset \realworld~\cite{Liao2022PAMI}. All the datasets are sequentially-collected sensory streams. Meanwhile, we develop a novel metric to measure the heading diversity of the retrieval results. In summary, our contributions are:
 
\begin{enumerate}
    \item We propose a novel self-supervised VPR framework termed \titlevariable~that eliminates pose-dependent supervision by iteratively refining pseudo labels from temporal and feature neighborhoods.
    \item We propose a new evaluation metric to assess the heading diversity of VPR retrieval results.
    \item We conduct comprehensive experiments using both RGB and point cloud datasets to show the auto-labeling and generalization capabilities of TF-VPR in comparison with other baseline methods.
    
\end{enumerate}





%% file: parts/2-related.tex
\section{Related Work}
\textbf{Visual place recognition}. Visual place recognition (VPR) is the problem of identifying a previously visited place based on visual information~\cite{zaffar2021vpr}. 
Existing Visual Place Recognition (VPR) methods mainly lie in two categories: (1) VPR techniques based on local features~\cite{cummins2011appearance,costante2013transfer,yuan2024local,sarlin2020superglue}, and (2) state-of-the-art VPR techniques using deep learning~\cite{arandjelovic2016netvlad,sunderhauf2015place,lopez2017appearance,naseer2017semantics,garg2018lost}. Within category (1), VPR techniques based on local features can be further subdivided according to the type of local features they utilize: hand-crafted or learnable. Hand-crafted approaches extract key points and descriptors manually, relying on local structures, gradient orientations, or intensity patterns in the image~\cite{cummins2011appearance,costante2013transfer}. In contrast, learnable feature approaches leverage machine learning algorithms to extract features, allowing for more flexible and adaptive representations~\cite{yuan2024local,sarlin2020superglue}. However, challenges such as perceptual aliasing and variations in view direction persist in this category~\cite{zaffar2021vpr,sheng2021nyu}. Within category (2), NetVLAD~\cite{arandjelovic2016netvlad} is a seminal deep-learning-based VPR framework, followed by various research extensions such as learning powerful feature representation~\cite{choy2019fully,deng2018ppfnet}, designing robust matching strategies~\cite{garg2022seqmatchnet,forechi2016sequential,vysotska2015lazy}, and investigating different input modalities in VPR~\cite{uy2018pointnetvlad,cai2021autoplace}. However, most methods are supervised by pose-dependent data~\cite{sunderhauf2015place,lopez2017appearance,arandjelovic2016netvlad}. 
To relax such a constraint, several attempts have been made~\cite{lowry2016supervised,merrill2018lightweight,kwon2021visual}, yet failed to handle diverse re-visiting viewpoints. 
The most relevant works to our method are VPR calibration{~\cite{lajoie2022self}} and the semi-parametric topological memory (SPTM){~\cite{savinov2018semi}}. To avoid direct supervision from GPS, the former approach{~\cite{lajoie2022self}} tries to obtain weak VPR supervision from SLAM. However, the quality of such supervision strongly depends on having good VPR for loop closing in SLAM in the first place, not to mention other challenges in SLAM. In contrast, the latter~\cite{savinov2018semi} utilizes temporal positives and negatives to train a binary classification network for adding edges for topological mapping, similar to VPR. However, since temporal neighbors tend to have very similar viewpoints, SPTM still struggles to recognize revisits of the same place from different viewpoints. To the best of our knowledge, no research has addressed \textit{self-supervised VPR} that can recognize places observed from \textit{various viewpoints} as \textit{either 2D images or 3D point clouds}.

\textbf{Noisy label learning.} Noisy labels become a problem as training data size increases, resulting in degraded performance~\cite{natarajan2013learning}.
To mitigate the issues of noisy labels, several learning attempts~\cite{sukhbaatar2014learning} have been made from directions including latent variable optimization~\cite{yu2018simultaneous}, loss function design~\cite{patrini2017making}, consistency Regularization~\cite{miyato2018virtual,hendrycks2019augmix}, and pseudo-label-based self-training~\cite{lee2013pseudo,reed2014training,zou2018unsupervised,zou2019confidence}. 


Among all pseudo-label methods, label refurbishment was first introduced by Bootstrapping~\cite{reed2014training}. Later on, another method addressed this problem using a self-training-based approach with an iterative workflow~\cite{zou2018unsupervised} which was used as a baseline architecture in their more recent framework confidence regularized~\cite{zou2019confidence}. This iterative approach has also been used in keypoint matching~\cite{yang2021self}. Our iterative training paradigm is similar at a high level but needs to address unique challenges and opportunities in obtaining image-level description for robust retrieval in VPR, unlike pixel-level segmentation~\cite{zou2018unsupervised,zou2019confidence} or keypoint-level descriptor matching~\cite{yang2021self} that cannot leverage temporal information.


\textbf{Contrastive learning.} It is a self-supervised learning technique to find feature representations that differentiate similar data pairs from dissimilar ones without labels. Data augmentation is often used, and the learning objective is to decrease the feature distances between the original and augmented images (positive samples), while increasing those distances between different images (negative samples)~\cite{he2020momentum,oord2018representation,tian2020contrastive}. In VPR, NetVLAD~\cite{uy2018pointnetvlad} uses the triplet loss which is similar to contrastive learning, yet relies on ground truth pose to define positive/negative samples.

\textbf{VPR evaluation.} There are several evaluation metrics for visual place recognition, \textit{e.g.}, the popular AUC-PR~\cite{lowry2015visual} provides a good overview of precision and recall performance but is less indicative in the cases when ground truth match could take multiple values. Recall Rate@N, as used in \cite{arandjelovic2014visual,torii201524,arandjelovic2016netvlad}, is designed to address such cases that the correct retrieval may be in the top-N results, and multiple correct query matches are neither penalized nor rewarded. However, existing VPR metrics rarely evaluate the viewpoint diversity of the retrieved results~\cite{zaffar2021vpr}, which is important in downstream applications such as SLAM. In this work, we develop such a metric to fill this gap. It assesses a VPR model's capacity to recognize places revisited from different directions. 

%% file: parts/3-method.tex
\providecommand{\mo}{\mathbf{o}}
\providecommand{\mq}{\mathbf{q}}
\newcommand{\lc}{\left ( }
\newcommand{\rc}{\right ) }

\providecommand{\positiveset}{\mathcal{P}_{{\bf{\mq}}_i}}
\providecommand{\postemporal}{\mathcal{P}^t_{{\bf{\mq}}_i}}
\providecommand{\negtemporal}{\mathcal{N}^t_{{\bf{\mq}}_i}}
\providecommand{\negset}{\mathcal{N}_{{\bf{\mq}}_i}}
\providecommand{\potpos}{\mathcal{\tilde{P}}_{{\bf{\mq}}_i}}
\providecommand{\verpos}{\hat{\mathcal{P}}_{{\bf{\mq}}_i}}
\providecommand{\feature}{f_{\theta}}

\section{Method}\label{sec:method}
\begin{figure*}[t]
    \centering {\includegraphics[trim={1.4cm 20.8cm 1.4cm 0.5cm},clip,width=0.98\textwidth
    ]{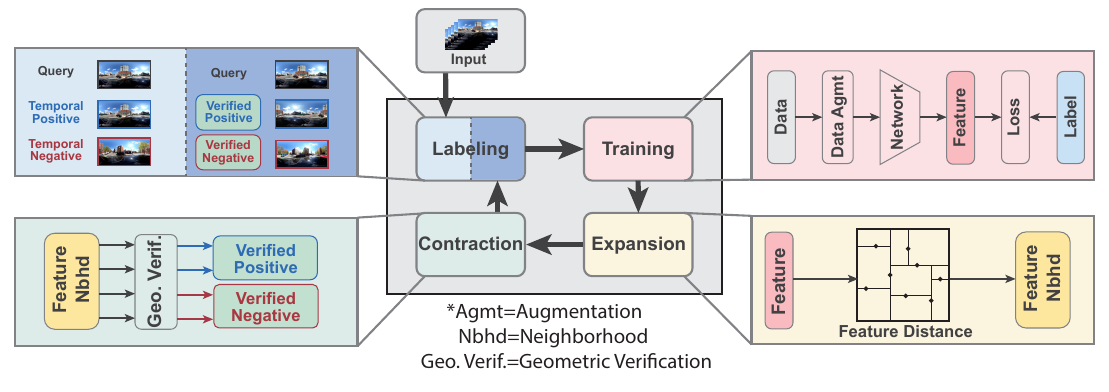}} \quad
    \caption{\textbf{Overview of \titlevariable.} A novel iterative method is designed for mining all query's spatial neighbors. Labeling, training, expansion, and contraction are four major steps in our approach. Labels can be refurbished  by iteratively learning feature representation, adding extra verified feature positives, and eliminating false positives to achieve self-supervised VPR.}
    \label{fig:workflow}
\vspace{-5mm}
\end{figure*}

\subsection{Problem setup and formulation}







Consider a mobile agent (e.g., a robot, a self-driving car) equipped with a panoramic RGB camera or a 360$^{\circ}$ LiDAR sensor while moving in an environment (typically in GPS-denied regions) without robust localization for accurate positions and orientations. At each step $i$ of its movement, the agent takes one observation $\bf{\mo}_i$. 
For cameras, $\bf{\mo}_i \in \mathbb{R}^{{H}\times {W}\times{3}}$ is an RGB image. For LiDARs, $\bf{\mo}_{i} = \{{\bf{p}}_{i}^j\}_{j=1}^{N}$ is a 3D point cloud, where $\bf{p}_i^j\in \mathbb{R}^{3}$ is the $j$-th point in the point cloud, and $N$ denotes the number of points.

To obtain a good spatial understanding of the environment, the agent needs visual or LiDAR place recognition to recognize revisits of places. Nowadays this is typically done by training a neural network $f$ with learnable parameters $\theta$ that can extract a global feature vector $\feature(\bf{\mo}_i)$ for each observation $\bf{\mo}_i$ in a training dataset, such that each observation's feature space neighbors are also its spatial neighbors. Our approach to addressing the task involves two primary modes: (1) \textit{generalization} and (2) \textit{auto-labeling}.

In \textit{generalization} mode, the evaluation criteria focus on the model's capacity to extend beyond the observed data to make predictions on previously unseen instances. The dataset $\mathbf{D}$ consists of two subsets: the training set $\mathbf{D}_{\text{train}}$ and the test set $\mathbf{D}_{\text{test}}$. During the training phase, the model $f_{\theta}$ is trained on a tuple $(\mathbf{q}, \mathbf{p}^{\mathbf{q}}, \mathbf{n}^{\mathbf{q}})$, where the query $\mathbf{q}$, its positives $\mathbf{p}^{\mathbf{q}}$, and its negatives $\mathbf{n}^{\mathbf{q}}$ are all drawn from the training set $\mathbf{D}_{\text{train}}$. Subsequently, during the testing phase, we partition the testing set $\mathbf{D}_{\text{test}}$ into the query set $\mathbf{Q}$ and the database set $\mathbf{D}_{\text{base}}$. The trained model $f_{\theta}$ is tested on all frames in $\mathbf{Q}$ and $\mathbf{D}_{\text{base}}$ to derive their feature vectors. Following this, for each query $\mathbf{q} \in \mathbf{Q}$, its top $k$ neighbors  are retrieved from the database $\mathbf{D}_{\text{base}}$ based on the euclidean feature distance $d_{f_{\theta}}^{2}(i,j)$ between the query and the database. These neighbors are compared with the ground truth to evaluate the model's cross-domain performance during inference.

The \textit{auto-labeling} mode's setup is mostly similar to  \textit{generalization}, except that \textit{auto-labeling} mode does not require a testing phase. Auto-labeling aims to generate a binary topology graph for a dataset $\mathbf{D}$. Each node in this graph represents an observation in $\mathbf{D}$, and an edge signifies that two observations are taken in proximity. Unlike the \textit{generalization} mode, we assume the ground truth of this graph is not available during auto-labeling (but only during the evaluation). In auto-labeling, we optimize the model $\feature$ on dataset D. Subsequently, a binary graph is generated based on the feature vectors extracted from $\feature$. An edge between two images can be connected if the feature distance $d_{f_{\theta}}^{2}(i,j)$ between them is below a threshold. Thus, auto-labeling is a self-supervised overfitting on the dataset $\mathbf{D}$. This process can generate topological graphs for SLAM, such as in DeepMapping2~\cite{chen2023deepmapping2}. 

Regardless of modes, the core idea of \titlevariable~is an iterative process of noisy label learning and refinement, designed to train a VPR network without relying on ground truth spatial neighborhoods for direct or weak supervision. This is achieved by utilizing the relationship between temporal, spatial, and feature neighborhoods. In Fig.~\ref{fig:workflow}, 
the pipeline consists of four stages discussed in the following subsections: (1) labeling, (2) training, (3) expansion, and (4) contraction.

\subsection{Initial data labeling and data augmentation}\label{Stage1} 
\begin{figure}[t]
    \vspace{2mm}
    \centering {\includegraphics[trim={0.1cm 0.0cm 0.6cm 0.6cm},clip,width=0.44\textwidth]{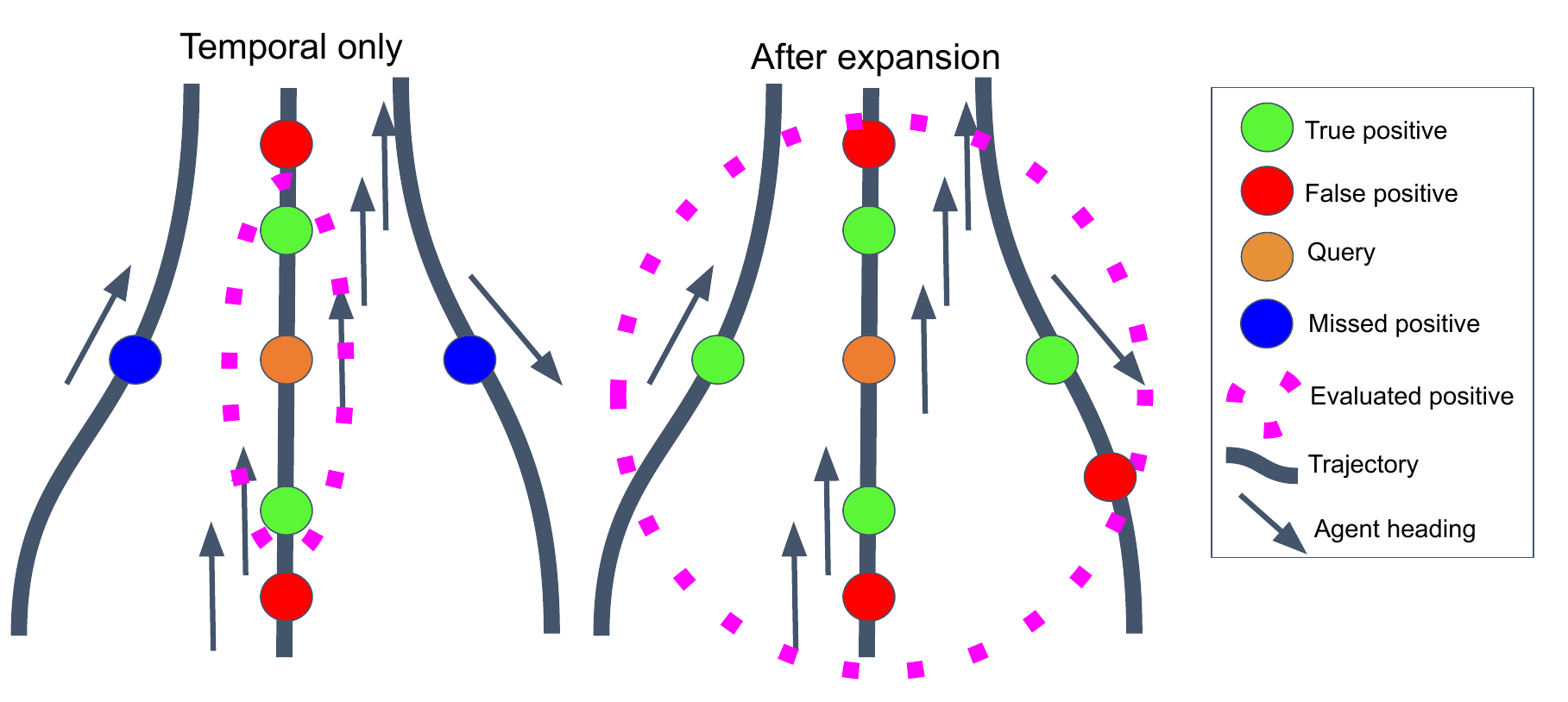}} \quad
    \vspace{-1mm}
    \caption{\textbf{Need of the expansion step.} On the left, training the network with only temporal neighbors limits evaluated positives to those with the same orientations as the query, missing spatial neighbors with different headings. Instead, on the right, combining data augmentation with iterative feature neighborhood expansion discover more spatial neighbors.}
    \label{fig:Search_alg}
\vspace{-6mm}
\end{figure}
Inspired from \cite{savinov2018semi,arandjelovic2016netvlad}, the temporal adjacent neighbor should be also spatially adjacent. In Fig.~\ref{fig:Search_alg} on the left, given a training query $\bf{q}_i$, we generate temporal positive set $\postemporal$ and negative set $\negtemporal$ based on the temporal index $i$.
\vspace{-0.2mm}
\begin{equation}\label{eq:1}
 {\bf{\mo}}_{j} \in \postemporal \Leftrightarrow|i-j|<n ,
\end{equation}
\vspace{-3mm}
\begin{equation}\label{eq:2}
{\bf{\mo}}_{j} \in \negtemporal \Leftrightarrow|i-j|>kn. 
\end{equation}

For a query $\mathbf{q}_i$, we select $n$ temporal neighbors ahead and behind it as its temporal positives, as indicated in Eq.~\eqref{eq:1}. Similarly, in Eq.~\eqref{eq:2}, its temporal negatives are defined as observations that are at least $kn$ time steps away. The hyperparameter $k$ in Eq.~\eqref{eq:2} helps to adjust the boundary for the temporal negatives.


Before feeding the data into the network for training, the augmentation module performs a random horizontal rolling on panoramic image input or a random rotation on point cloud input. The augmentation step simulates a random rotation of the sensor heading, allowing the network to obtain new observations from different orientations at the same location. This is critical if the same location is visited multiple times from different viewpoints. Following forward propagation of the network, we compute the loss and perform back propagation using the pseudo label from the first stage.

\textbf{Discussion on the initial labels}: The current setup may contain noisy labels, but in practice, their proportion is very low. This is because empirically the ratio of positives to negatives for each query is low, making the probability of sampling spatial positives from temporal negatives very low.

\subsection{Expansion}\label{Stage2}

After Sec.\ref{Stage1}, expanding the neighborhoods is necessary because the current positive set $\positiveset$ can be limited in size and diversity. Additionally, the negative set $\negset$ may be noisy since temporally distant observations could still be spatial neighbors due to place revisits, as shown in Fig.~\ref{fig:Search_alg}.


Between epochs, our iterative workflow expands neighborhoods by evaluating the training dataset, as the features from the network encode the spatial information. For each query $\mathbf{\mq}_i$, we apply KNN on feature space to retrieve the top \textit{k} feature neighbors as potential positives $\potpos$. The number of retrieved neighbors is determined dynamically since not all queries have the same number of positives. Specifically,
we first compute the minimum feature distance $d_{f_{\theta}}^{2}$ between each query $\bf{q}_i$ and its temporal neighbors $\postemporal$ as threshold 

\vspace{-2mm}
\begin{equation}\label{eq:3}
\tau_i=\min\limits_{\mo_i\in\postemporal} \left( d_{f_{\theta}}^{2}\left(\bf{q}_i, \bf{\mo}_i\right)\right).
\end{equation}

Next, for the sake of computation speed, we only focus on finding the positive candidates from the top \textit{k} nearest feature neighbors using KNN. An observation $\mo_j$ in these feature neighbors, with a smaller feature distance than the threshold $\tau_i$, will be selected as a potential candidate




\vspace{-2mm}
\begin{equation}\label{eq:4}
\potpos=\left\{\bf{\mo}_j : d_{f_{\theta}}^{2} \left(\bf{q}_i, \bf{\mo}_j\right)<\tau_i\right\}.
\end{equation}

Existing VPR training methods typically prioritize hard negative mining over positive mining, both utilizing the feature space to refine the training dataset. However, in Sec.~\ref{sec:experiment}, we emphasize the crucial role of positive mining. It reduces the performance gap between self-supervised and supervised methods by increasing dataset variety.

\subsection{Contraction}\label{CVM}
\vspace{-.5mm}

Geometric verification checks the validity of the feature neighborhoods $\potpos$ (or positive candidates from Sec.\ref{Stage2}) before merging elements in $\potpos$ into $\positiveset$.  Eq.~\eqref{eq:verification} describes how the verified positives $\verpos$ merges $\positiveset$ during each epoch's contraction step, where $e$ represents the epoch number. The false positive in $\potpos$ is noisy and might harm the learning process of the network. 

Depending on different input types, the verification step varies and isolates from the network. In this paper, we use RANSAC for image verification and ICP for point cloud verification. The way of contracting positives is similar to the expansion of positive candidates. Similar to Eq.\eqref{eq:3} and Eq.\eqref{eq:4}, we calculate the minimum matching score threshold $\epsilon_i$ using RANSAC or ICP between the query and its temporal neighbors. Subsequently, we assess the matching scores between all positive candidates $\potpos$ and the query. Any positive candidate with a matching score surpassing threshold $\epsilon_i$ is classified as verified. Thus, verified positives $\verpos$ are trustworthy and permanently added into the positive set $\positiveset$
\vspace{-1mm}
\begin{equation}\label{eq:verification}
\mathcal{P}^{(e)}_{{\bf{q}}_i} = \mathcal{P}^{(e-1)}_{{\bf{q}}_i}  \cup  \hat{\mathcal{P}}^{(e)}_{{\bf{q}}_i},
\end{equation}

\subsection{Loss function}
We use triplet loss $L_{\theta}$ for the $i$th training tuple $(\bf{q}_i, {{\bf{p}}^{\bf{q}}_{i}}, {{\bf{n}}^{\bf{q}}_{i}} )$ in our implementation as shown in Eq.~\eqref{eq:6}, where ${\bf{p}}^{\bf{q}}_{i} \in \positiveset$ and  ${\bf{n}}^{\bf{q}}_i\ \in \negset$. And $\positiveset$ and $\negset$ represent the positive and negative set updated from the previous iteration, $m$ is hyperparameter (set to 0.2 by default). This loss is used by~\cite{arandjelovic2016netvlad} to pull positive features together and repel negative features so that it will be easier to apply KNN and obtain the potential geographical nearest neighbors.
\vspace{-1mm}
\begin{equation}\label{eq:6}
L_{\theta} =\sum_{{\bf{q}}_i} l\left(\min _{i} d_{f_{\theta}}^{2}\left({\bf{q}}_i, {\bf{p}}^{\bf{q}}_{i}\right)+m-d_{f_{\theta}}^{2}\left({\bf{q}}_i, {\bf{n}}^{\bf{q}}_{i}\right)\right),
\end{equation}



%% file: parts/4-experiments.tex
\section{Experiments}\label{sec:experiment}

\titlevariable~ is tested in two kinds of environments: 
simulated RGB images~\cite{savva2019habitat}, and real-world point cloud~\cite{Liao2022PAMI}. Additional real-world RGB experiments are provided in the appendix due to space constraints. Our codebase uses PyTorch~\cite{pytorch} with network parameters optimized using Adam~\cite{kingma2014adam}. The learning rate is set to 1e-4 together with a weight decay of 1.0e-7. We compare \titlevariable~with both supervised and self-supervised baseline methods. 

\textbf{Notation}:  We conducted a complete ablation study on both datasets. \textit{Any method incorporating any of the following modules should be regarded as TF-VPR.} In the experiment section, we specifically denote TF-VPR as SPTM+A+F for convenience in the ablation study. Each module discussed in Sec.{~\ref{sec:method}} is abbreviated as follows:
\begin{enumerate}
    \item ``+A" denotes the method with data augmentation.
    \item ``+F" denotes the method iteratively expanding spatial neighborhoods (including expansion and contraction) 
\end{enumerate}

\subsection{Evaluation metrics}\label{evaluate_metric}

\textbf{Recall rate (Recall@N)}\label{recall_N} measures the ratio of successful retrievals to the total number of queries. A retrieval is considered successful if at least one of the top-N retrieved results is a ground-truth spatial neighbor of the query.
Ground truth can be obtained by K-D tree search on geographical location $(x,y,z)$ within a certain radius $R$. It is important to exclude temporal neighbors of a query from its top-N retrievals when computing this metric. In our setup, temporal-based methods can easily overfit the temporal neighborhood, leading to uninformative evaluations with recall rates always at $100\%$ if temporal neighbors are kept in the ground truth.


\textbf{Heading diversity (\text{HD)}} measures the diversity of sensor headings of the true positives (w.r.t. the query's heading) among the top-$|GT|$ retrievals. $|GT|$ is the size of the ground-truth set based on a specific radius $R$ as described in Recall@N. Those hard positives with headings different from the query are more valuable for loop closure in downstream applications~\cite{chen2023deepmapping2}. Thus, to evaluate this heading diversity, we evenly divide the 360$^{\circ}$ range of headings into 8 angular bins. The first and last bins are excluded because they contain positives with similar headings w.r.t. the query. In this case, the $m$-th bin covers the heading difference range ${\mathcal{Q}_m}$ as:
\begin{equation}\label{eq:qud_def}
\mathcal{Q}_m = [m\times45^{\circ}, (m+1)\times45^{\circ}],\,m\in[1,2,3,4,5,6].
\end{equation}

\begin{figure}[t]
    \vspace{2mm}
  \begin{center}
    \includegraphics[width=0.40\textwidth]{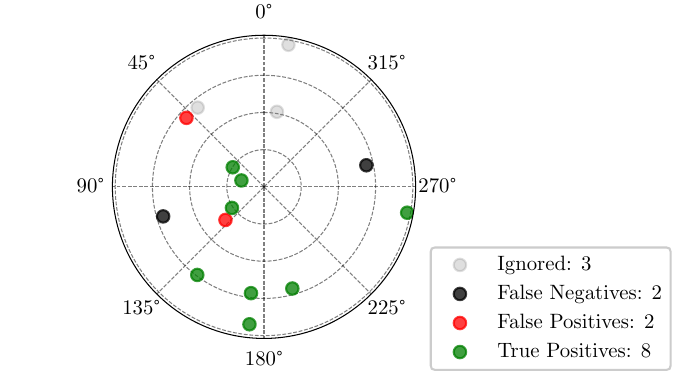}
    \caption{\textbf{Heading diversity illustration}. The angle represents the heading difference between the query and the evaluated positives. HD represents how many angular bins are covered by true positives vs. that by the ground truth. The figure gives an example of how to calculate HD. Excluding the first and last bins, $\potpos$ contains 10 retrieved non-temporal positives, 8 of which are true positives, and they fall into 5 different bins, while ground truth covers 6 bins, so HD is $5/6$.}
    \label{fig:HD_metric}
  \end{center}
  \vspace{-6.5mm}
\end{figure}

Then, we define HD for a query $\bf{q}$ as the bin coverage ratio between the true positives and the ground truth:
\begin{equation}\label{eq:HD_N}
\text{HD(\bf{q})}=\frac{\sum_{m\in{[1...6]}}\mathds{1}(\exists \bf{x}\in\tilde{\mathcal{P}}^{|GT|}_{{\bf{\mq}}} \land (\theta_{\bf{q}}-\theta_{\bf{x}}) \in \mathcal{Q}_m)}{\varepsilon + \sum_{m\in{[1...6]}}\mathds{1}(\exists \bf{y}\in\mathcal{P}^{|GT|}_{{\bf{\mq}}} \land (\theta_{\bf{q}}-\theta_{\bf{y}}) \in \mathcal{Q}_m)},
\end{equation}
where $\theta_{\bf{q}}$ and $\theta_{\bf{x}}$ are respectively the heading of the query and a frame ${\bf{x}}$, $\tilde{\mathcal{P}}^{|GT|}_{{\bf{\mq}}}$ is the set of true positives in the top-$|GT|$ retrievals, $\mathcal{P}^{|GT|}_{{\bf{\mq}}}$ is the ground truth positive set, $\varepsilon$ is an arbitrarily small positive quantity to avoid zero division error, and $\mathds{1}(\cdot)$ is the indicator function. See Fig.~\ref{fig:HD_metric} for an example. Finally, we report the average HD for all queries. 


\subsection{Experiments on \realworld~dataset}\label{sec:realrgb}
\begin{table}[t]
\centering
\captionsetup{font=sc}
\captionsetup{font={scriptsize, sc, stretch=1.3}, justification=centering, labelsep=newline}
\vspace{2mm}
\caption{\textbf{Auto-labeling on KITTI-360 dataset}. We employ the best recall rate (R) and heading diversity (HD) to evaluate these training procedures. Experiments are conducted with an overfitting setup: both the training and testing sets are identical. We emphasize the top three performing baselines using {\textcolor{red}{red}}, {\textcolor{teal}{teal}}, and {\textcolor{blue}{blue}} highlights.}
\label{tab:kitti_al}
\vspace{1mm}
\resizebox{1\linewidth}{!}{
\begin{tabular}{cl|cccc|cccc}
\hline
\multicolumn{2}{c|}{Scene}                                              & \multicolumn{4}{c|}{Drive\_0000}                          & \multicolumn{4}{c}{Drive\_0005}                          \\ \hline
\multicolumn{2}{c|}{Metric}                                        & \multicolumn{1}{c|}{HD}     & R@1   & R@5    & R@10   & \multicolumn{1}{c|}{HD}     & R@1   & R@5    & R@10       \\ \hline
\multicolumn{2}{c|}
{SPTM~\cite{savinov2018semi}}                 & \multicolumn{1}{c|}{\textcolor{teal}{8.17}}        & \textcolor{teal}{82.04}  & \textcolor{teal}{89.22}  & \textcolor{teal}{92.15}  & \multicolumn{1}{c|}{9.29} & \textcolor{teal}{71.84}  & \textcolor{teal}{77.02} & \textcolor{teal}{80.58}  \\

\multicolumn{2}{c|}{SPTM+A(Ours)} & \multicolumn{1}{c|}{\textcolor{teal}{15.09}} &  78.92  & \textcolor{blue}{84.80}  & \textcolor{blue}{86.76}  & \multicolumn{1}{c|}{\textcolor{teal}{16.31}} & 64.40  & 72.17  & 75.73  \\

\multicolumn{2}{c|}{SPTM+F(Ours)}             & \multicolumn{1}{c|}{\textcolor{red}{16.42}}    & \textcolor{red}{82.35}  & \textcolor{red}{91.18}  & \textcolor{red}{93.63}  & \multicolumn{1}{c|}{\textcolor{red}{16.52}} & \textcolor{red}{72.82}  & \textcolor{red}{81.88}  & \textcolor{red}{83.17}  \\

\multicolumn{2}{c|}{SPTM+A+F(Ours)}        & \multicolumn{1}{c|}{\textcolor{blue}{14.77}}         & \textcolor{blue}{79.90}  & \textcolor{blue}{84.80}  & 85.78  & \multicolumn{1}{c|}{\textcolor{blue}{14.83}} & \textcolor{blue}{64.72}  & 71.84  & \textcolor{blue}{77.99}  \\\hline


\end{tabular}
}
\vspace{-6mm}
\end{table}

\begin{figure*}
\vspace{2mm}
\begin{subfigure}{0.5\textwidth}
        \includegraphics[width=\textwidth]{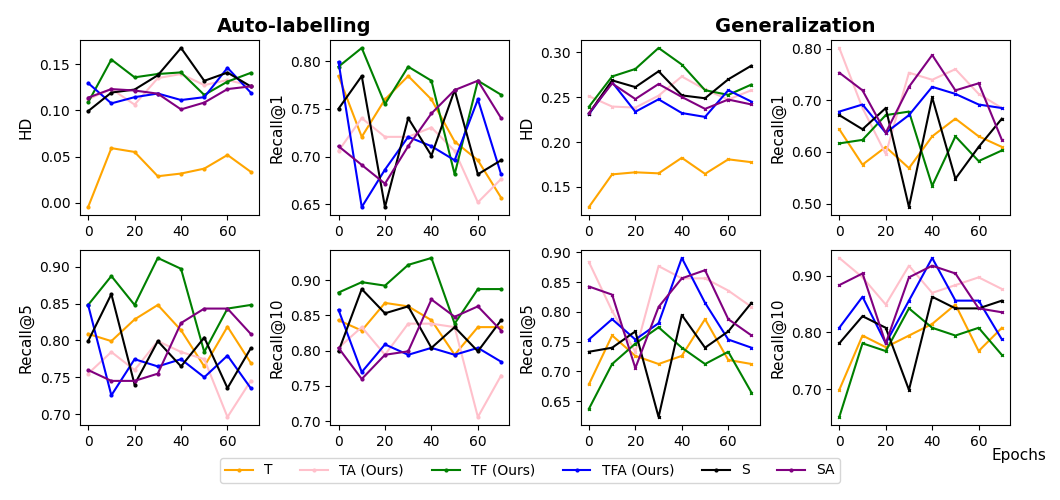}

    \end{subfigure}
    \begin{subfigure}{0.5\textwidth}
        \includegraphics[width=\textwidth]{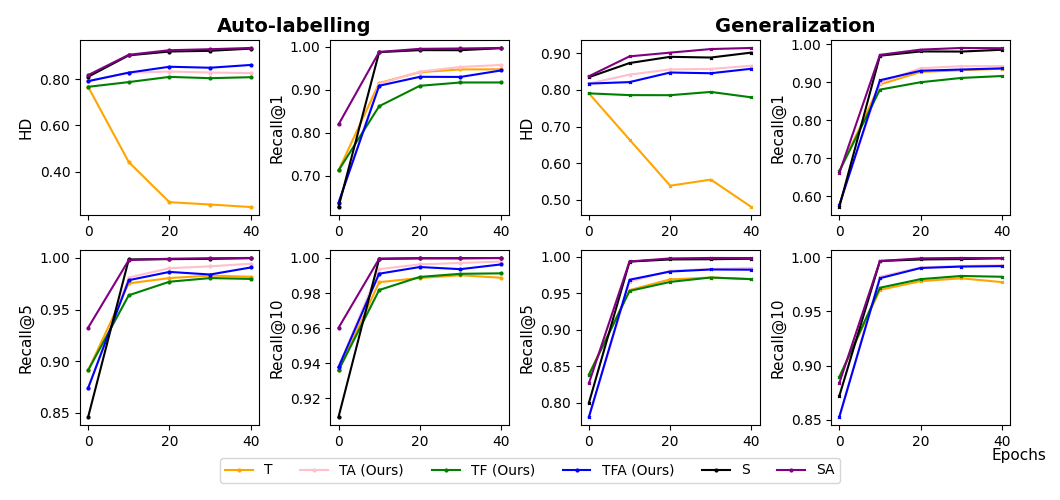}

    \end{subfigure}
\vspace{-6mm}
   \caption{Heading Diversity, Recall@1, Recall@5, and Recall@10 versus training epochs on the scene Drive\_0000/ 00.02 from the KITTI-360 dataset(left) and the scene Reyno/R.G from Habitat-Sim dataset(right). Abbreviations: T denotes SPTM{~\cite{savinov2018semi}}. S denotes the supervised method{~\cite{uy2018pointnetvlad}}. A and F follows the notation in Sec.{~\ref{sec:experiment}}.}
\label{fig:realworld_habitats}
\vspace{-4mm}
\end{figure*}

\textbf{Dataset.} 
KITTI-360~\cite{Liao2022PAMI} dataset is an outdoor point cloud dataset captured with a Velodyne HDL-64E LiDAR and a SICK LMS 200 laser scanner in a roof-mounted pushbroom setup. This dataset comprises 100,000 laser scans covering a driving distance of 73.7 kilometers. We select three scenes for our study: Drive\_0000 and Drive\_0005 for training, and Drive\_0002 for testing, containing 33,110 LiDAR scans. After pre-processing, we downsample each frame to 10,000 points, forming point clouds of size (10000, 4). Additionally, the GPS for each frame is provided for validation.

\textbf{Baseline methods.} The following baselines are evaluated: (1) PointNetVLAD~\cite{uy2018pointnetvlad} trained with pose-based supervision, (2) PointNetVLAD trained with temporal pseudo labels (SPTM)~\cite{savinov2018semi}, (3) prototypical contrastive learning (PCL)~\cite{li2020prototypical} as another self-supervised VPR method used in visual navigation~\cite{kwon2021visual}, and (4) the ablation study baselines. 

\textbf{Implementation details.} 
Our implementation is based on PointNetVLAD. For training, we select $\delta = 5$ and $k = 2$ in Eq.~\ref{eq:1} and Eq.~\ref{eq:2}, and sample 8 pairs of positives and negatives from the candidate pool for each query. For evaluation, we exclude the closest $30$ temporal neighbors for each query from the top-N retrievals as explained in Sec.~\ref{evaluate_metric}.

One baseline PCL is implemented with its default settings, tuning the total number of clusters. We set the number of clusters to $200$, $500$, and $1000$, and select the best result. 


\textbf{Auto-labeling on~\realworld.} From Table{~\ref{tab:kitti_al}}, SPTM+F shows superior performance in auto-labeling, outperforming SPTM with a 2\% increase in recall rate and an 8\% improvement in heading diversity. This improvement is due to the infrequent revisits of scenes from different angles in typical outdoor settings. Consequently, the addition of augmentation imposes an \textit{extra burden} on the network, resulting in a decrease in both recall and heading diversity (HD).


\textbf{Generalization on~\realworld.}
SPTM+A and TF-VPR (SPTM+A+F) demonstrate strong generalization capabilities on the \realworld~dataset, as shown in Fig.~\ref{fig:realworld_habitats}. This improvement is due to the increased diversity in the training dataset through augmentation.
It also quantitatively demonstrates a similar improvement in both recall rate and heading diversity for SPTM+A and~\titlevariable. Furthermore, Table~\ref{tab:kitti_tt} demonstrates that~\titlevariable~outperforms other baselines in both scenes, achieving approximately $4\%$ and $12\%$ higher scores in terms of recall rate and heading diversity, respectively.

\textbf{Compare~\titlevariable~with supervised methods.} From Table~\ref{tab:kitti_tt}, we also observe that SPTM+A and \titlevariable~comparable or even superior performance compared to the supervised method trained on the same dataset (small domain gap), significantly outperforming the one trained on the Oxford robot car dataset (large domain gap)~\cite{maddern20171}. This performance gap is due to the substantial domain differences between the training and testing datasets. Note that TF-VPR is a self-supervised method that allows us to collect data from similar scenes for self-supervised training. In contrast, other supervised baselines are often constrained by the availability of ground truth data (data privacy, sensor availability, annotation costs, and etc.). Therefore, when the domain gap between training and testing sets is small, TF-VPR data training holds an \textbf{inherent advantage}.

 \begin{table}[t]
\centering
\captionsetup{font=sc}
\captionsetup{font={scriptsize, sc, stretch=1.3}, justification=centering, labelsep=newline}

\caption{\textbf{Generalization on KITTI-360 dataset}. We evaluate the generalization capability of methods based on their best recall rate (Recall) and heading diversity (HD). Experiments are categorized into three groups. The first group is cross-validation, involving all methods except the last two rows (e.g., 00.02 indicates training on Drive\_0000 and testing on Drive\_0002). The second group, marked with $*$, involves inference-only, where we use pretrained models on the test set (refer to the footnote). The last group, represented by the last row, is a supervised one with a small domain gap, where both the training and test sets are from two scenes of the same dataset. This group theoretically establishes an upper bound for all self-supervised methods. Other notation follows Table{~\ref{tab:kitti_al}}. \textbf{Abbreviations: sdg. (small domain gap), ldg. (large domain gap).}} 
\label{tab:kitti_tt}
\vspace{-1mm}
\resizebox{1\linewidth}{!}{
\begin{tabular}{cl|cccc|cccc}
\hline
\multicolumn{2}{c|}{Scene}                                              & \multicolumn{4}{c|}{00.02.}                          & \multicolumn{4}{c}{05.02.}                          \\ \hline
\multicolumn{2}{c|}{Metric}                                 & \multicolumn{1}{c|}{HD}              & R@1   & R@5    & R@10    & \multicolumn{1}{c|}{HD}   & R@1   & R@5    & R@10      \\ \hline
\multicolumn{2}{c|}{PCL~\cite{li2020prototypical}}                & \multicolumn{1}{c|}{0.55}        & 42.58  & 53.13  & 70.13  & \multicolumn{1}{c|}{12.11} & 51.11  & 63.32  & 70.98  \\

\multicolumn{2}{c|}
{SPTM~\cite{savinov2018semi}}               & \multicolumn{1}{c|}{20.23}           & 71.92  & \textcolor{blue}{86.30}  & \textcolor{blue}{90.41} & \multicolumn{1}{c|}{\textcolor{blue}{21.27}}  & \textcolor{blue}{80.30}  & \textcolor{teal}{90.15}  & \textcolor{blue}{93.84}  \\

\multicolumn{2}{c|}{SPTM+A(Ours)}  & \multicolumn{1}{c|}{\textcolor{blue}{29.37}} &  \textcolor{red}{81.51}  & \textcolor{teal}{89.03}  & \textcolor{teal}{93.14} & \multicolumn{1}{c|}{\textcolor{red}{35.10}}  & \textcolor{red}{82.19}  & \textcolor{red}{91.78}  & \textcolor{red}{95.89} \\

\multicolumn{2}{c|}{SPTM+F(Ours)}  & \multicolumn{1}{c|}{\textcolor{red}{34.69}} & \textcolor{blue}{72.60}  & 82.19  & 87.67 & \multicolumn{1}{c|}{17.23}  & \textcolor{teal}{80.88} & 88.23  & 90.69 \\

\multicolumn{2}{c|}{SPTM+A+F(Ours)}  & \multicolumn{1}{c|}{\textcolor{teal}{31.62}} & \textcolor{teal}{80.82}  & \textcolor{red}{89.04}  & \textcolor{red}{93.15}  & \multicolumn{1}{c|}{\textcolor{teal}{33.76}}  & 78.76  & \textcolor{blue}{89.73}  & \textcolor{teal}{93.85} \\\hline

\multicolumn{2}{c|}{Supervised\_ldg.$*${~\cite{uy2018pointnetvlad}}}           & \multicolumn{1}{c|}{39.58}     & 19.95 & 25.06 & 27.11 & \multicolumn{1}{c|}{N/A} & N/A & N/A & N/A \\\hline

\multicolumn{2}{c|}{Supervised\_sdg.{~\cite{uy2018pointnetvlad}}}          & \multicolumn{1}{c|}{33.53}     & 75.34 & 91.78 & 93.83 & \multicolumn{1}{c|}{29.77}  & 86.99 & 96.58 & 97.94 \\

\hline
\end{tabular}
}
\begin{tablenotes}
\small 
    \item \textit{$*$ represents the supervised PointNetVLAD method pretrained on the Oxford RobotCar dataset~\cite{maddern20171} and used for inference on the scene Drive\_0002. }
\end{tablenotes}

\vspace{-5mm}
\end{table}



\subsection{Experiments on Habitat-Sim dataset}\label{sec:RGB_SIM}

\begin{table*}[t]
\centering
\captionsetup{font=sc}
\captionsetup{font={scriptsize, sc, stretch=1.3}, justification=centering, labelsep=newline}
\vspace{2mm}
\caption{\textbf{Auto-labeling on Habitat-Sim dataset}. Avg. represents the average metric of all scenes. The metrics, colors, and abbreviations follow Tab.{\ref{tab:kitti_al}}.}
\label{tab:habitat_al}
\resizebox{1.0\linewidth}{!}{
\begin{tabular}{l|llll|llllllllllll}
\hline
Metrics                                                      & \multicolumn{4}{l|}{HD}                   & \multicolumn{4}{l}{Recall@1}                     & \multicolumn{4}{l}{Recall@5}                     & \multicolumn{4}{l}{Recall@10}                                              \\ \hline
Datasets                                                                          & Goffs           & Reyno           & Stilwell     & Avg.      & Goffs           & Reyno           & Stilwell    & Avg.       & Goffs           & Reyno           & Stilwell      & Avg.     & Goffs          & Reyno          & Stilwell   & Avg.         \\ \hline
SPTM~\cite{savinov2018semi}              & 81.73          & 78.06          & \textcolor{blue}{79.77}        & 79.85          & 95.57          & 95.14          & 95.91  & 95.54        & 98.72          & 98.45          & 98.55     & 98.57     & 99.30          & 99.12          & 99.18       & 99.20            \\
SPTM+A(ours)                                                        & \textcolor{teal}{87.58}          & \textcolor{teal}{83.35}          & \textcolor{teal}{87.35}      &  \textcolor{teal}{86.09}       & \textcolor{teal}{96.77}          & \textcolor{red}{96.24}          & \textcolor{teal}{98.23}   &  \textcolor{teal}{97.08} & \textcolor{red}{99.59}          & \textcolor{red}{99.51}          & \textcolor{blue}{99.71} & \textcolor{red}{99.60} & \textcolor{teal}{99.82}          & \textcolor{red}{99.85}          & \textcolor{blue}{99.91} & \textcolor{red}{99.86}\\
SPTM+F(ours)                                                                    & \textcolor{blue}{85.94}          & \textcolor{blue}{82.04}          & 79.22  & \textcolor{blue}{82.40}
& \textcolor{blue}{96.51}          & \textcolor{teal}{95.96}          & \textcolor{blue}{98.01}          &  \textcolor{blue}{96.82}
& \textcolor{blue}{99.36}          & \textcolor{blue}{98.53}          & \textcolor{teal}{99.92}          & \textcolor{blue}{99.27}
& \textcolor{teal}{99.82}          & \textcolor{blue}{99.40}          & \textcolor{teal}{99.94}          & \textcolor{blue}{99.72}        \\
SPTM+A+F(ours)                                                         & \textcolor{red}{88.64} & \textcolor{red}{86.51}  & \textcolor{red}{89.99}      & \textcolor{red}{88.38}     & \textcolor{red}{96.99} & \textcolor{blue}{95.68} & \textcolor{red}{98.82} & \textcolor{red}{97.16}  & \textcolor{teal}{99.44} & \textcolor{teal}{99.08} & \textcolor{red}{99.77}     & \textcolor{teal}{99.42}     & \textcolor{red}{99.89} & \textcolor{teal}{99.63} & \textcolor{red}{99.97}    & \textcolor{teal}{99.83}                \\ \hline
\end{tabular}}
\vspace{-2mm}
\end{table*}

\begin{table*}[t]
\centering
\captionsetup{font=sc}
\captionsetup{font={scriptsize, sc, stretch=1.3}, justification=centering, labelsep=newline}

\caption{\textbf{Generalization on Habitat-Sim dataset}. G.S. indicates the model is trained on the scene Goffs and is tested on the scene Stilwell, and R. denotes the scene Reyno. The groups, colors, and abbreviations follow Tab.{\ref{tab:kitti_tt}} and Tab.{\ref{tab:habitat_al}}. \textbf{Abbreviations: sdg. (small domain gap), ldg. (large domain gap), G(Goffs),S (Stilwell), R(Reyno).}}
\label{tab:habitat_tt}
\vspace{-1mm}
\resizebox{1.0\linewidth}{!}{
\begin{tabular}{l|llll|llllllllllll}
\hline
Metrics                                                     & \multicolumn{4}{l|}{HD}                     & \multicolumn{4}{l}{Recall@1}                     & \multicolumn{4}{l}{Recall@5}                     & \multicolumn{4}{l}{Recall@10}                                             \\ \hline
Datasets                                                                        & G.S.           & R.G.           & S.R.       & Avg.    & G.S.           & R.G.           & S.R.  & Avg.          & G.S.           & R.G.           & S.R.    & Avg.       & G.S.           & R.G.           & S.R.      & Avg.     \\ \hline
PCL~\cite{li2020prototypical}   & 81.16          & 73.66          & 75.51         & 76.78                  & 25.16          & 24.82          & 29.95      & 26.64    & 43.54          & 41.99          & 49.46  & 45.00         & 52.63          & 51.49          & 57.50      & 53.87             \\

SPTM~\cite{savinov2018semi}                            & 82.27          & \textcolor{blue}{82.85}          & \textcolor{blue}{79.74}   & 81.62    & \textcolor{blue}{94.94}          & \textcolor{teal}{94.82}          & 90.63     & \textcolor{blue}{93.46}    & 97.84          & \textcolor{blue}{97.81}         & 95.76    & 97.14      & 98.71          & 98.60          & 97.02 & 98.11               \\
SPTM+A(ours)                                                     & \textcolor{teal}{87.63}          & \textcolor{teal}{87.81}          & \textcolor{teal}{81.85}     & \textcolor{teal}{85.76}          & \textcolor{teal}{95.50}          & \textcolor{blue}{94.60}          & \textcolor{blue}{92.36} 
&  \textcolor{teal}{94.15}& 
\textcolor{red}{99.01}          & \textcolor{teal}{98.94}         & \textcolor{blue}{97.75}  &  \textcolor{teal}{98.57}  & \textcolor{teal}{99.54}          & \textcolor{teal}{99.58}          & \textcolor{teal}{98.76}  & \textcolor{teal}{99.29}\\
SPTM+F(ours)                                                       & \textcolor{blue}{83.94}          & 82.39          & 79.35      & \textcolor{blue}{81.89}        & 94.48         & 92.59          & \textcolor{teal}{92.73}  & 93.27        & \textcolor{blue}{98.26}          & 97.63          & \textcolor{teal}{98.02}   & \textcolor{blue}{97.97}       & \textcolor{blue}{99.12}          & 98.76         & 98.75  & \textcolor{blue}{98.88}                \\
SPTM+A+F(ours)                                                         & \textcolor{red}{87.75} & \textcolor{red}{88.51} & \textcolor{red}{82.09} & \textcolor{red}{86.12}               & \textcolor{red}{96.15} & \textcolor{red}{95.30} & \textcolor{red}{93.93} & \textcolor{red}{95.13}  & \textcolor{teal}{98.39} & \textcolor{red}{98.95} & \textcolor{red}{98.42}   & \textcolor{red}{98.59}      & \textcolor{red}{99.68} & \textcolor{red}{99.62} & \textcolor{red}{99.16}   & \textcolor{red}{99.49}            \\ \hline

Supervised\_ldg.$*$~\cite{arandjelovic2016netvlad}  &   58.14       &   62.01        &    67.67    & 62.61     &    22.68         &      35.21     &   22.68   &  26.86    &     32.72      &    47.05      &    32.72       &  37.50   &  36.93       &      51.01     &   36.93  & 41.62            \\

CNNI.R.$\triangle$~\cite{radenovic2018fine}                    & 73.73         & 71.87         & 70.84   & 72.15         & 91.75          & 93.31          & 91.66      & 92.24    & 98.10         & 98.34          & 97.78      & 98.07   & 98.82          & \textcolor{blue}{99.21}     &  \textcolor{blue}{98.78}   & 98.94              \\\hline

VLAD+SIFT~\cite{arandjelovic2013all,jegou2010aggregating}
   & 82.80          & 84.00          & 84.46 & 83.75 & 79.91          & 88.24          & 86.51 & 84.89         & 87.53          & 93.77          & 92.45  & 91.25        & 90.56          & 95.59          & 93.82       & 93.32         \\
Supervised\_sdg.~\cite{arandjelovic2016netvlad}            & 92.41          & 90.53          & 85.82     & 89.59   & 99.47          & 98.83          & 97.67      & 98.66    & 99.94          & 99.78          & 99.56      & 99.76    & 99.98          & 99.93          & 99.86   & 99.92           \\\hline

\end{tabular}}
\begin{tablenotes}
\small
    \item \textit{$*$ represents that we employ the supervised NetVLAD method pretrained on the model pretrained on Pittsburgh dataset~\cite{torii2013visual}.} \textit{$\triangle$ represents CNNI.R.{~\cite{radenovic2018fine}} which is re-designed based on the VPR task. The model used is pretrained on multiple datasets~\cite {radenovic2018fine}. We test both methods on the scene Goffs, Stilwell, and Reyno.}
\end{tablenotes}
\vspace{-4.5mm}
\end{table*}

\textbf{Dataset.}
\titlevariable~is also tested via Habitat-Sim~\cite{savva2019habitat} simulator on the Gibson photorealistic RGB dataset~\cite{xiazamirhe2018gibsonenv}, which offers panoramic RGB images for a variety of indoor scans. We capture RGB images with a panoramic camera mounted on a robot moving randomly in the virtual environment, resulting in a total of $33,679$ RGB images across three Gibson rooms. Each image is downsampled to $256 \times 64$ pixels. In contrast to other datasets, this simulated RGB dataset contains a large number of revisits of places from both similar and different directions, which is useful for testing recall rate and heading diversity for VPR.

\textbf{Baseline methods.}
Following Sec.~\ref{sec:realrgb}, we use SPTM~\cite{savinov2018semi}, NetVLAD~\cite{arandjelovic2016netvlad}, VLAD~\cite{arandjelovic2013all,jegou2010aggregating}, and PCL~\cite{li2020prototypical} as baselines. Additionally, we introduce the CNN image retrieval baseline CNNI.R.{~\cite{radenovic2018fine}} and a classic non-deep-learning VPR method VLAD{~\cite{jegou2010aggregating}{~\cite{arandjelovic2013all}} for comparison.}

\textbf{Implementation details.}
We set $\delta = 5$ and $k = 2$ in Eq.~\ref{eq:1} and Eq.~\ref{eq:2}. For each query, we exclude its closest $30$ temporal neighbors from the top-N retrievals as explained in \ref{evaluate_metric}. For the VLAD baseline, we use $128$-dimensional SIFT features with a cluster size of $128$. The raw VLAD descriptor dimension of $128 \times 128$ is reduced to $512$ by PCA. Besides, CNNI.R. uses pretrained model to extract the feature and follows the previous implementation for evaluation.

\textbf{PCL, VLAD, CNNI.R., and conventional visual SLAM system.} Table~\ref{tab:habitat_tt} shows the poor performance of PCL. It may not be ideal for VPR solutions because most contrastive learning methods tend to form disjoint clusters for each category, which may not effectively represent continuous features in vision-based SLAM problems. Contrarily, VLAD performs effectively as we aggregate features and evaluate performance on the same dataset, akin to network overfitting. However, the evaluation process is time-consuming compared to our learnable model, which only requires inference to a new model. Despite this, our method outperforms VLAD by approximately $8\%$ in terms of recall rate. Furthermore, CNNI.R., an image retrieval method, lacks the precision of most VPR methods and exhibits slightly inferior performance compared to SPTM. It also lags noticeably behind our proposed approach, particularly in terms of recall@1, where it trails by approximately $4\%$. Finally, we tested the conventional visual SLAM system, like OpenVSLAM{~\cite{sumikura2019openvslam}}. However, OpenVSLAM easily loses track, with a total of $17.75\%$ of frames lost during the tracking. OpenVSLAM constructs a disjoint topology graph while tracking odometry, resulting in a recall rate of $54.98\%$ compared to $93.93\%$ for TF-VPR. Due to its poor performance, we did not consider OpenVSLAM as a benchmark.

\textbf{Auto-labeling on Habitat-Sim.} Table{~\ref{tab:habitat_al}} shows that TF-VPR, SPTM+A, and SPTM+F achieve the highest auto-labeling recall rate, with TF-VPR outperforming SPTM by about $1\%$. Similar to Sec.~\ref{sec:realrgb}, feature neighborhoods(F) plays a critical role play in overfitting the test dataset. However, unlike the discussion in Sec.~\ref{sec:realrgb}, augmentation enhances recall due to the abundant viewpoint variation and lack of lighting change in the simulated RGB dataset. Besides, TF-VPR outperforms all self-supervised baselines in heading diversity, with an improvement of around $8\%$.

\textbf{Generalization on Habitat-Sim.} 
In Table{~\ref{tab:habitat_tt}}, TF-VPR outperforms all self-supervised baselines and approaches the supervised NetVLAD performance and VLAD+SIFT in both recall rate and heading diversity (HD). Besides, TF-VPR improves recall rate by $3\%$ and heading diversity by $5\%$. 

\textbf{Compare TF-VPR with supervised methods.}. Similar to Sec.~\ref{sec:realrgb}, Table{~\ref{tab:habitat_tt}} indicates that the domain gap challenges the supervised method. SPTM+A and TF-VPR are comparable to the supervised method trained on the same dataset (small domain gap), outperforming the one trained on the Pittsburgh dataset (large domain gap)~\cite{maddern20171}. Additional per-frame performance visualizations are in the appendix.

\textbf{Adapbility of TF-VPR on other backbones.} We tested TF-VPR's adaptability with backbones beyond NetVLAD and PointNetVLAD, showing significant improvements in heading diversity and recall rate. Table{~\ref{tab:diff_bb}} shows that TF-VPR effectively applies to various backbones (MixVPR)~\cite{ali2023mixvpr} and PADLoC~\cite{arce2023padloc}) and modalities (point cloud and RGB), with each version outperforming its counterparts. This versatility enhances results across diverse scenarios.

\begin{table}[t]
\centering
\captionsetup{font=sc}
\captionsetup{font={scriptsize, sc, stretch=1.3}, justification=centering, labelsep=newline}

\caption{\textbf{Autolabelling on the point cloud (HABITAT-SIM) and RGB (KITTI) datasets. We evaluate the autolabelling capability of TF-VPR against two other backbones: MixVPR{~\cite{ali2023mixvpr}} and PADLoC{~\cite{arce2023padloc}}. The implementation details adhere to the original codebase and we use the pretrained models provided by the original authors for the baselines. Other abbreviations follow Table}{~\ref{tab:kitti_al}}}.
\label{tab:diff_bb}
\vspace{-1mm}
\resizebox{1.0\linewidth}{!}{
\begin{tabular}{c|l|cccc}
\hline
Data Type                   & Metric    & HD    & R@1   & R@5   & R@10  \\ \hline
\multirow{2}{*}{RGB}        & MixVPR    & 63.47 & 48.01 & 79.66 & 86.50 \\
                            & TF-VPR\_M* & 70.98 & 53.91 & 82.74 & 89.46 \\ \hline
\multirow{2}{*}{Pointcloud} & PADLoC    & 25.42 & 39.47 & 44.71 & 46.34 \\
                            & TF-VPR\_P* & 35.42 & 43.73 & 49.22 & 51.19 \\ \hline
\end{tabular}
}
\begin{tablenotes}
\small 
    \item \textit{$*$ M represents TF-VPR model with MixVPR backbone and P represents TF-VPR model with PADLoC backbone. TF-VPR model is trained for $5$ epochs.}
\end{tablenotes}

\vspace{-5mm}
\end{table}

%% file: parts/5-conclusions.tex
\vspace{1mm}
\section{Conclusion}\label{sec:conclusion}
We propose \titlevariable~as a self-supervised VPR method adaptive to both auto-labeling and generalization tasks for determining the unknown spatial neighbors from the fixed temporal neighbors and learnable feature neighbors. While our method primarily utilizes panoramic RGB images and 360-degree point clouds, this iterative nature of TF-VPR allows for easy extension to non-panoramic inputs. Extensive experiments show that \titlevariable~not only improves the recall rate over the existing method but also can retrieve spatial positives with more diverse viewpoints on various datasets. \titlevariable~enables easier use of VPR in real-world robotics and computer vision applications, and can be applied to most existing deep-learning-based VPR methods.

\textbf{Limitations.} The bottlenecks of TF-VPR models lie in their generalization across large domain gaps. Specifically, one prevalent challenge in VPR is the impact of appearance changes, arising from factors such as lighting variations, weather, or seasonal transitions. TF-VPR shows limitations in generalization in these contexts and is more suitable for autolabeling tasks and VPR within the same scene.



%% file: parts/Appendix.tex
\renewcommand{\thetable}{\Roman{table}}
\renewcommand{\thefigure}{\Roman{figure}}
\renewcommand\thesection{\Roman {section}}

\section*{Appendix}

\setcounter{section}{0}
\setcounter{figure}{0}
\setcounter{table}{0}
\providecommand{\titlevariable}{TF-VPR}

In the supplementary material, we provide additional visualizations and analyses including: (1) computation time of various methods during the training and inference phases; (2) qualitative visualization of autolabelling results; (3) auto-labelling test on our self-collected real-world RGB dataset(NYU-VPR-360); and (4) dataset visualizations, with ground truth trajectory.

\section{Computation Time}

We report the computation time in the training and inference phases in Table~\ref{tab:time}, formatted as $hh:mm:ss$. Regarding the training phase, the computation includes network optimization and label refurbishment. Compared with other hand-crafted or learnable methods, TF-VPR for autolabelling tasks only needs to consume more time for training, but does not introduce an extra burden for inference. The training process is more time-consuming due to feature neighborhoods, which require positive expansion as well as pairwise geometric verification. Note that compared to brute-force pairwise geometric verification, our training is more affordable thanks to the elaborate temporal and feature neighborhoods expansion.

\vspace{-2mm}
\begin{table}[h]
\centering
\captionsetup{font=sc}
\captionsetup{font={scriptsize, sc, stretch=1.3}, justification=centering, labelsep=newline}
\caption{\textbf{Training and inference time} of each method on three different datasets. Train represents the training time for one epoch. Eval. represents inference time. Pair. Verif. means pairwise verification.}
\label{tab:time}
\vspace{-1mm}
\resizebox{1\linewidth}{!}{
\begin{tabular}{|c|cc|cccccc|clcllllclllll|}
\hline
\multirow{2}{*}{Dataset} & \multicolumn{2}{c|}{Point Cloud} & \multicolumn{6}{c|}{Habitat-Sim Scene }  & \multicolumn{13}{c|}{Real RGB} \\
\cline{2-22}   & \multicolumn{2}{c|}{Scene} & \multicolumn{2}{c|}{Goffs}   & \multicolumn{2}{c|}{Micanopy}  & \multicolumn{2}{c|}{Spotswood}   & \multicolumn{2}{c|}{Scene 1}   & \multicolumn{11}{c|}{Scene 2}    \\ \hline
Metric                   & \multicolumn{1}{c|}{Train}        & Eval.       & \multicolumn{1}{c|}{Train} & \multicolumn{1}{c|}{Eval.} & \multicolumn{1}{c|}{Train} & \multicolumn{1}{c|}{Eval.} & \multicolumn{1}{c|}{Train} & Eval. & \multicolumn{1}{c|}{Train}     & \multicolumn{1}{l|}{Eval.} & \multicolumn{5}{c|}{Train}     & \multicolumn{6}{c|}{Eval.}  \\ \hline
SPTM                     & \multicolumn{1}{c|}{0:49:14}             &    0:00:12            &  \multicolumn{1}{c|}{0:58:34}      & \multicolumn{1}{c|}{0:13:23}         & \multicolumn{1}{c|}{0:34:57}      & \multicolumn{1}{c|}{0:07:29}         & \multicolumn{1}{c|}{1:28:12}      &   0:09:13       & \multicolumn{1}{c|}{2:21:36}          & \multicolumn{1}{l|}{0:05:59}         & \multicolumn{5}{c|}{0:40:04}          & \multicolumn{6}{c|}{0:14:33}          \\
PCL                      & \multicolumn{1}{c|}{N/A}          &   N/A             & \multicolumn{1}{c|}{0:15:45}      & \multicolumn{1}{c|}{0:01:20}         & \multicolumn{1}{c|}{0:17:32}      & \multicolumn{1}{c|}{0:01:15}         & \multicolumn{1}{c|}{0:19:24}      &      0:01:12    & \multicolumn{1}{c|}{0:28:55}          & \multicolumn{1}{l|}{0:09:29}         & \multicolumn{5}{c|}{0:13:25}          & \multicolumn{6}{c|}{0:07:19}          \\
TF-VPR                     & \multicolumn{1}{c|}{5:38:12}             &    0:00:06            & \multicolumn{1}{c|}{21:01:06}      & \multicolumn{1}{c|}{0:09:46}         & \multicolumn{1}{c|}{12:38:49}      &  \multicolumn{1}{c|}{0:05:33}         & \multicolumn{1}{c|}{15:03:50}      &  0:06:28        & \multicolumn{1}{c|}{3:33:46} & \multicolumn{1}{l|}{0:04:16}         & \multicolumn{5}{c|}{8:35:10} & \multicolumn{6}{c|}{0:09:54} \\ \hline
VLAD                     & \multicolumn{1}{c|}{N/A}          &  N/A              & \multicolumn{1}{c|}{0:02:14}      & \multicolumn{1}{c|}{0:02:39}         & \multicolumn{1}{c|}{0:05:29}      & \multicolumn{1}{c|}{0:04:31}         & \multicolumn{1}{c|}{0:03:46}      &  0:04:09        & \multicolumn{1}{c|}{0:12:07}          & \multicolumn{1}{l|}{2:44:50}         & \multicolumn{5}{c|}{0:14:19}          & \multicolumn{6}{c|}{0:52:09}          \\
NetVLAD               & \multicolumn{1}{c|}{3:57:09}             &      0:00:27          & \multicolumn{1}{c|}{0:54:04}      & \multicolumn{1}{c|}{0:06:42}         & \multicolumn{1}{c|}{:0:31:27}      & \multicolumn{1}{c|}{0:04:17}         & \multicolumn{1}{c|}{0:40:08}      &    0:05:01      & \multicolumn{1}{c|}{1:00:06}          & \multicolumn{1}{l|}{0:14:41}         & \multicolumn{5}{c|}{0:22:41}          & \multicolumn{6}{c|}{0:04:16}          \\ \hline
Pair. Verif.  & \multicolumn{2}{c|}{46731:25:31} & \multicolumn{2}{c|}{6630:16:19}   & \multicolumn{2}{c|}{2294:31:07}  & \multicolumn{2}{c|}{2915:44:43}   & \multicolumn{2}{c|}{11083:40:57}   & \multicolumn{11}{c|}{5350:30:50}         \\ \hline
\end{tabular}
}   
\label{Time_table}
\end{table}



\vspace{-5mm}

\section{Autolabelling visualization} 
\begin{figure}
\begin{center}
{\includegraphics[width=0.5\textwidth
    ]{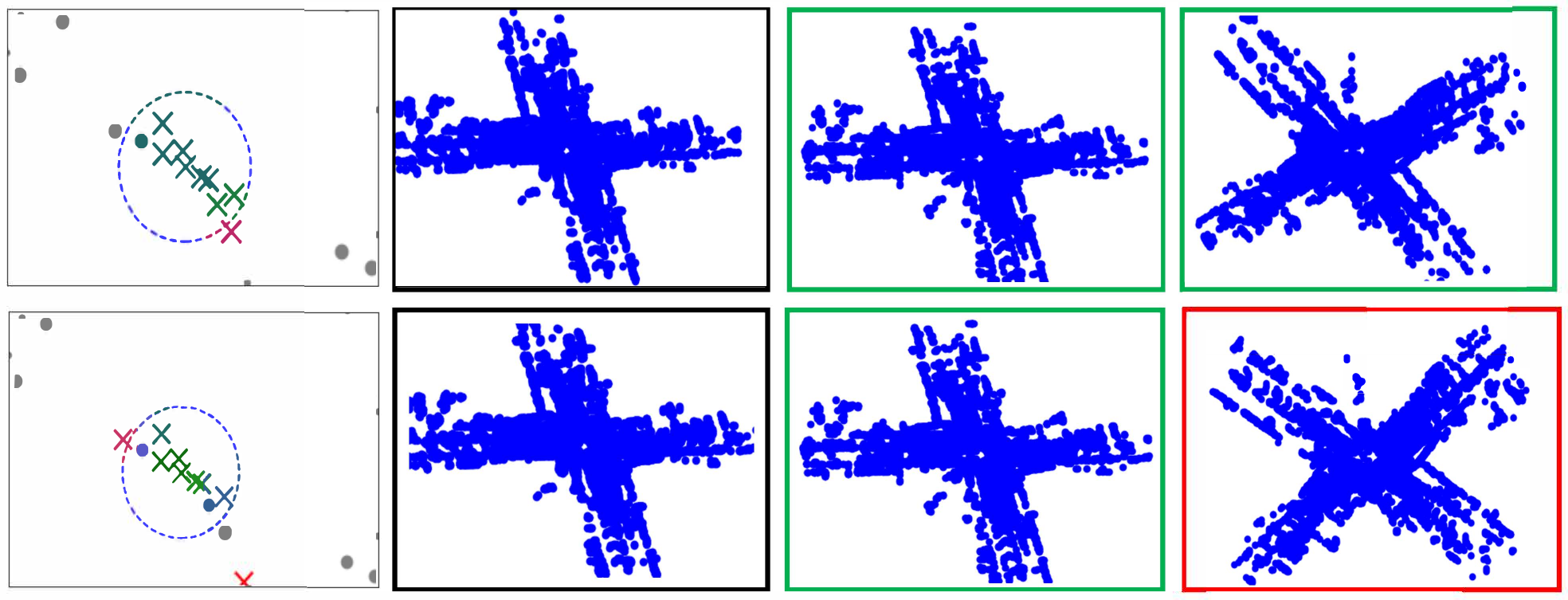}}
\end{center}
\vspace{-1mm}
   \caption{\textbf{Top 10 positive retrievals on KITTI-360 dataset}. In the upper row, \textbf{TF-VPR} excels in accurately retrieving neighbors from different directions.  In contrast, the lower row demonstrates instances where SPTM may include false positives in its top 10 retrievals. From left to right, the columns represent 1. scatter map of top 10 retrievals where  {\textcolor{black}{black}}, {\textcolor{teal}{green}} and {\textcolor{red}{red}} points represent query, true-positives, and false-positives;  2. the query frame; 3-4. two samples from the top 10 retrievals. Frames are considered positives only if they are within $2$ meters of the query.}
   
\label{fig:kitti_qual}
\vspace{-2mm}
\end{figure}

\begin{figure}
\begin{center}
{\includegraphics[width=0.5\textwidth
    ]{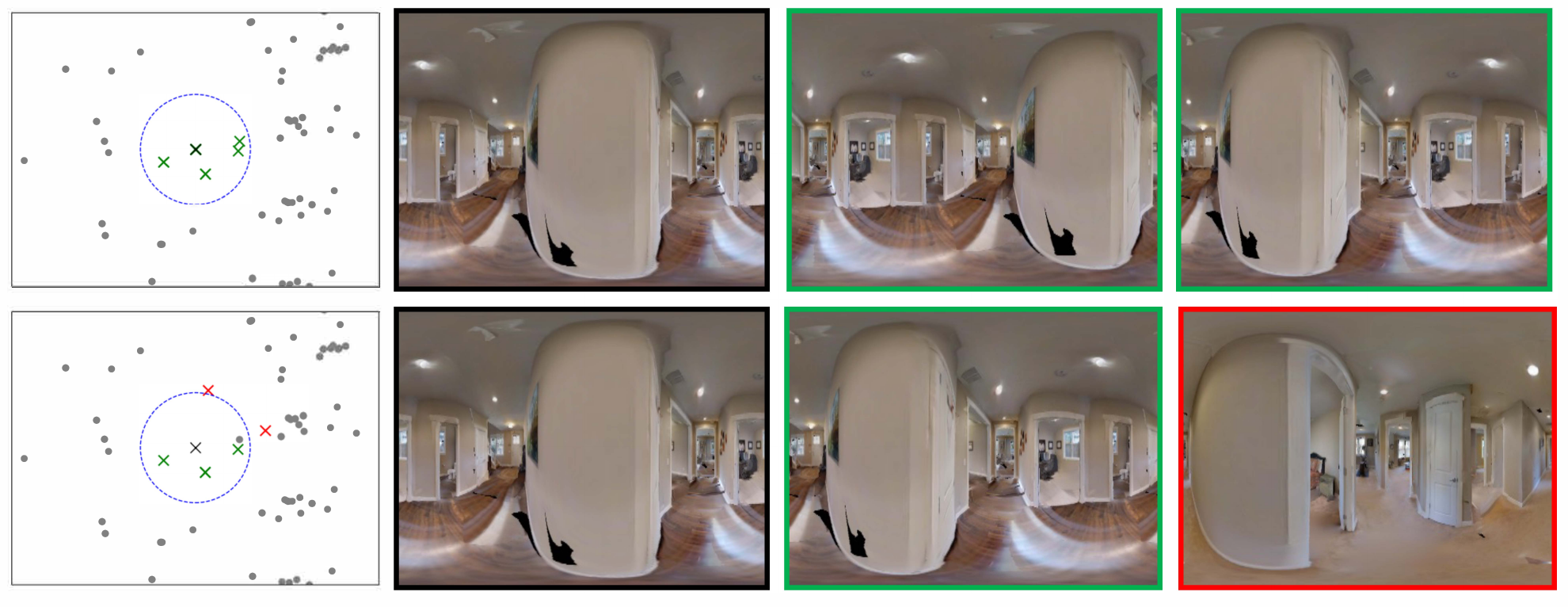}}
\end{center}
\vspace{-1mm}
\caption{\textbf{Top 10 positive retrievals on Habitat-Sim dataset.}
All image interpretations follow Fig.{~\ref{fig:kitti_qual}}.}
\label{fig:qual_habitat}
\vspace{-5mm}
\end{figure}

In this section, we provide further qualitative illustrations of the experimental results. For the KITTI-360 dataset, we provide the query point cloud, its predicted positives, and the scatter map representing their locations as shown in Fig.{~\ref{fig:kitti_qual}}. We demonstrate that TF-VPR exhibits stronger retrieval capabilities.  In contrast, SPTM fails in recall@10 and HD estimates. Similarly, Fig.{~\ref{fig:qual_habitat}} also demonstrates an improvement in the retrieval quality on the Habitat-Sim dataset. Augmentation and feature neighborhoods enrich the training dataset, enabling more accurate spatial neighbor retrievals on unseen data during inference.

\section{Autolabelling on real-world RGB dataset}

\textbf{NYU-VPR-360 dataset.} Although several VPR datasets feature panoramic images~\cite{torii2013pittsburgh,torii201524}, few include multiple visits to the same location from various angles. To demonstrate our method's capability in retrieving images with different headings, we introduced the NYU-VPR-360 dataset, captured using a GoPro MAX (a dual-lens 360$^{\circ}$ camera with GPS recording). This dataset consists of sequentially collected panoramic RGB images of street views in New York City, with the GoPro mounted on the roof of a driving vehicle.
We utilize GPS readings from the camera to provide ground truth for spatial neighborhoods. We select key panoramic images from the video footage and ensure that they are accurately synchronized with the GPS data. The dataset is composed of two driving trajectories, covering an area of approximately $80,000 m^2$. There are over $15,000$ images with a resolution of $3840 \times 1920$ pixels, each accompanied by its corresponding location for each scene. Most junctions in the dataset allow at least two types of maneuvers (such as turning left or right) with different driving directions, facilitating VPR from various angles and perspectives. However, a few intersections are restricted due to traffic regulations. Fig.{~\ref{fig:real-rgb}} shows some examples of images and robot trajectories.

\textbf{Baselines.} Following Sec.~\ref{sec:RGB_SIM}, we use SPTM~\cite{savinov2018semi}, NetVLAD~\cite{arandjelovic2016netvlad}, VLAD~\cite{arandjelovic2013all,jegou2010aggregating}, and PCL~\cite{li2020prototypical} as baselines. 

\textbf{Implementation details.} 
The images are resized to $128 \times 64$ pixels. We set $n = 10$ and $u = 5$ as outlined in Sec.~\ref{Stage1}. For Scene 1 and Scene 2, due to the varying sampling rates of the sensors, we exclude the closest $30$ and $100$ temporal neighbors, respectively, from the top-N retrievals for each query, as explained in Sec.~\ref{evaluate_metric}.

\begin{figure}[]
    \centering
    \includegraphics[trim={0.75cm 0.2cm 0.8cm 0}, clip, width=0.48\textwidth]{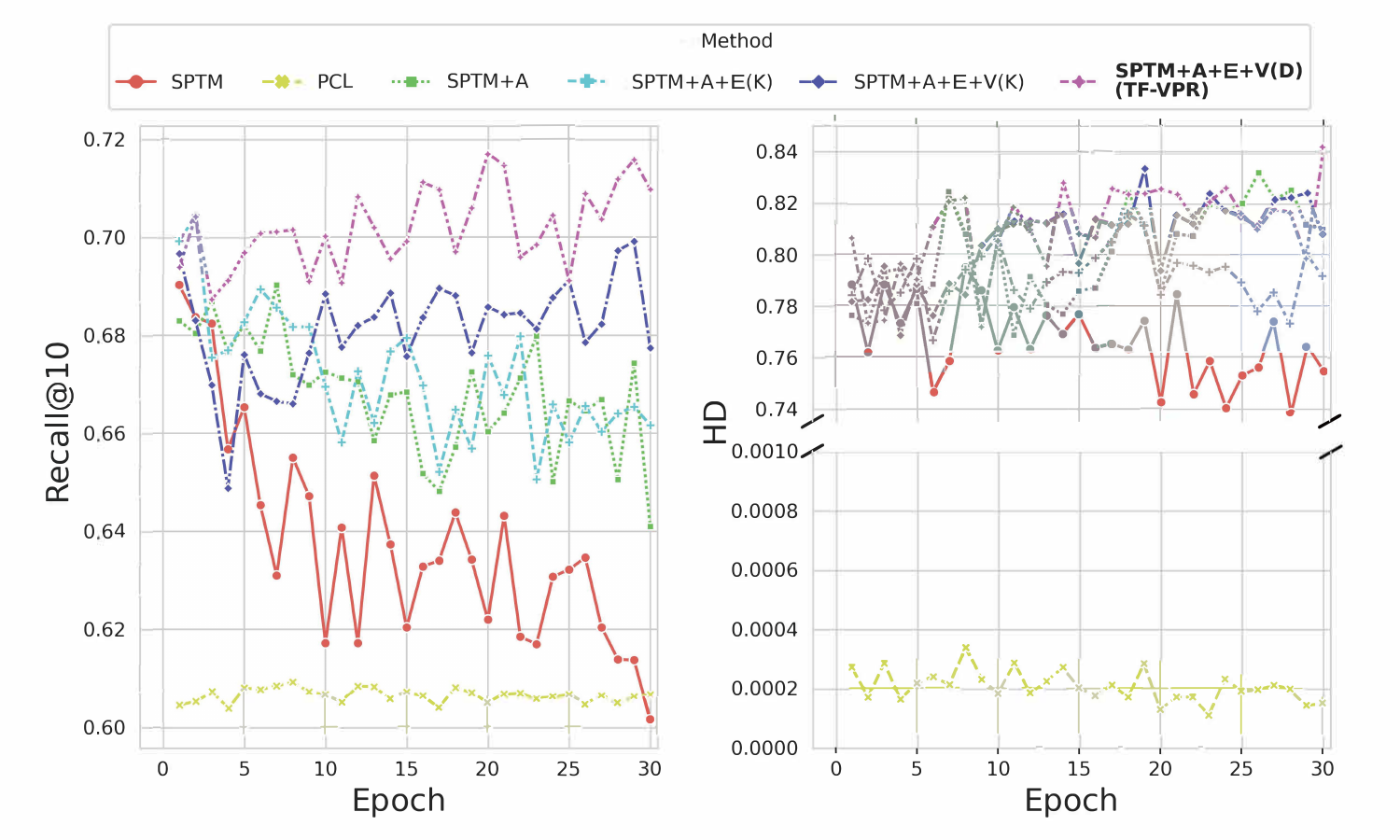}
    \caption{\textbf{Recall@10 and HD vs. training epoch} on NYU-VPR-360 (Scene 1). We conducted a thorough comparison in both the expansion (E) and contraction (V) steps. $K$ represents the method with KNN expansion, which uses a fixed number of frames during its corresponding step. In contrast, $D$ denotes the per-frame threshold methods utilized by TF-VPR. Other abbreviations follow Fig.{~\ref{fig:realworld_habitats}}.}
    \label{fig:realrgb_exp}
    \vspace{-5mm}
\end{figure}

\textbf{Per-frame threshold mechanism.} In both Sec.~\ref{Stage2} and Sec.~\ref{CVM}, we apply an expansion and contraction process using a per-frame threshold $\tau_i$. This approach, which calculates a threshold for each frame, allows for a more flexible selection and filtering of positive samples compared to methods that use a fixed number for expansion and contraction. By comparing against this adaptive threshold, we can better account for the varying contexts of each frame. For example, places like intersections naturally have larger spatial neighborhoods than hallways. To design this frame-specific mechanism, we propose the following hypothesis, as shown in Eq.\ref{eq:3} and Eq.\ref{eq:4}: if an observation is closer to the query than at least one of the query's temporal neighbors in the feature space, then that observation is more likely to be a ground truth spatial neighbor of the query. Consequently, we conducted a detailed ablation study, presented in Fig.~\ref{fig:realrgb_exp}, to compare our flexible approach with methods that select a fixed number of frames for expansion or contraction. We observe that incorporating contraction leads to better overall performance than using KNN for expansion alone. Furthermore, for contraction, the per-frame threshold method outperforms the KNN approach. In this work, E+V(D) is equivalent to F, which is highlighted in Sec.~\ref{sec:experiment}.

\begin{figure}[t]
    \centering {\includegraphics[width=0.44\textwidth
    ]{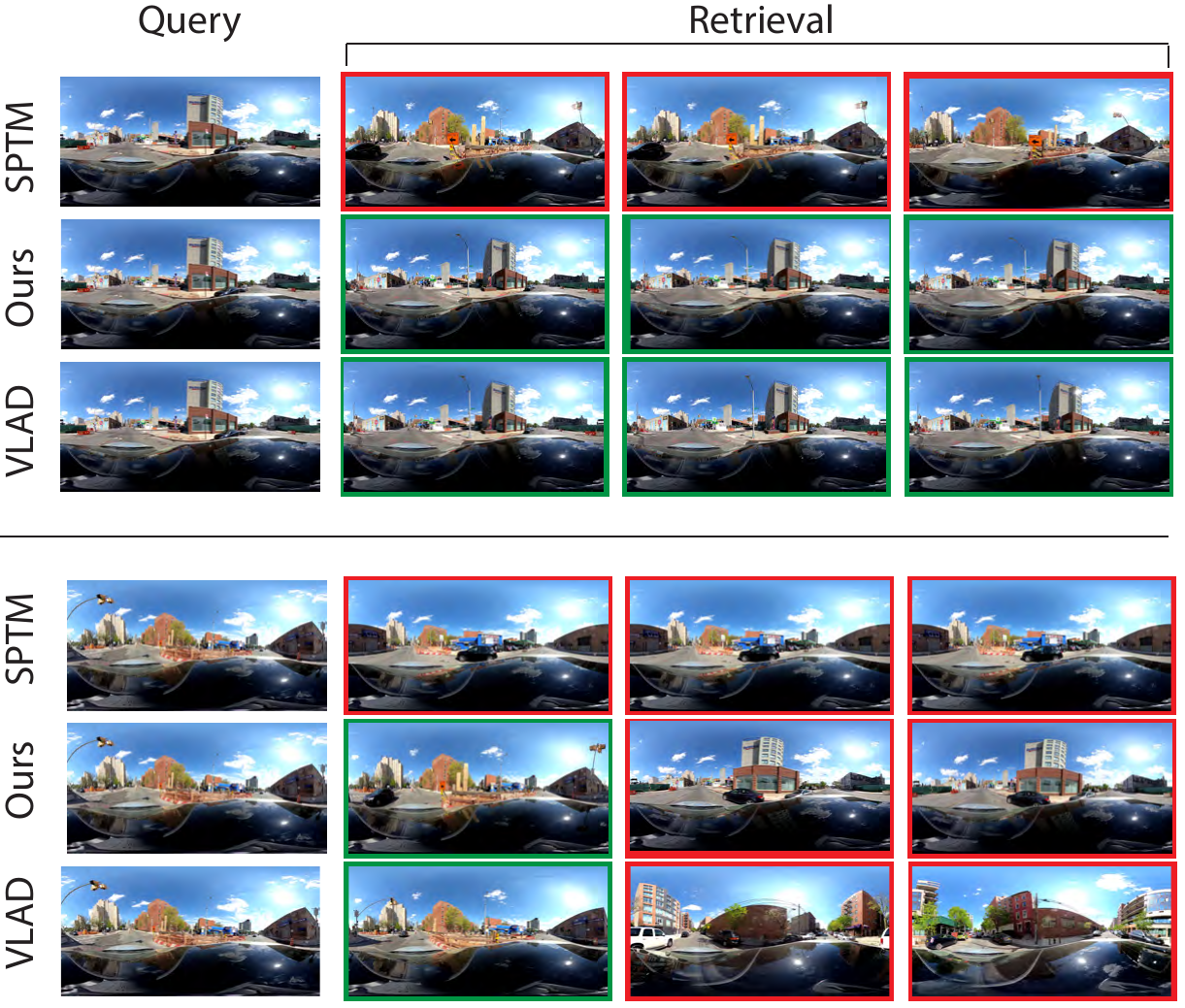}}
    \vspace{-2mm}
    \caption{\textbf{Qualitative VPR results on NYU-VPR-360 dataset}. The upper part shows an example where TF-VPR outperforms SPTM, and the lower part shows a challenging example in the dataset where only the top-1 retrieval of TF-VPR and VLAD is correct. {\textcolor{teal}{Green}} and {\textcolor{red}{red}} respectively indicate true and false positives.}
    \label{fig:query_realrgb}
\vspace{-10mm}
\end{figure}

\begin{figure}[t]
    \begin{center}
    \includegraphics[width=1.0\linewidth]{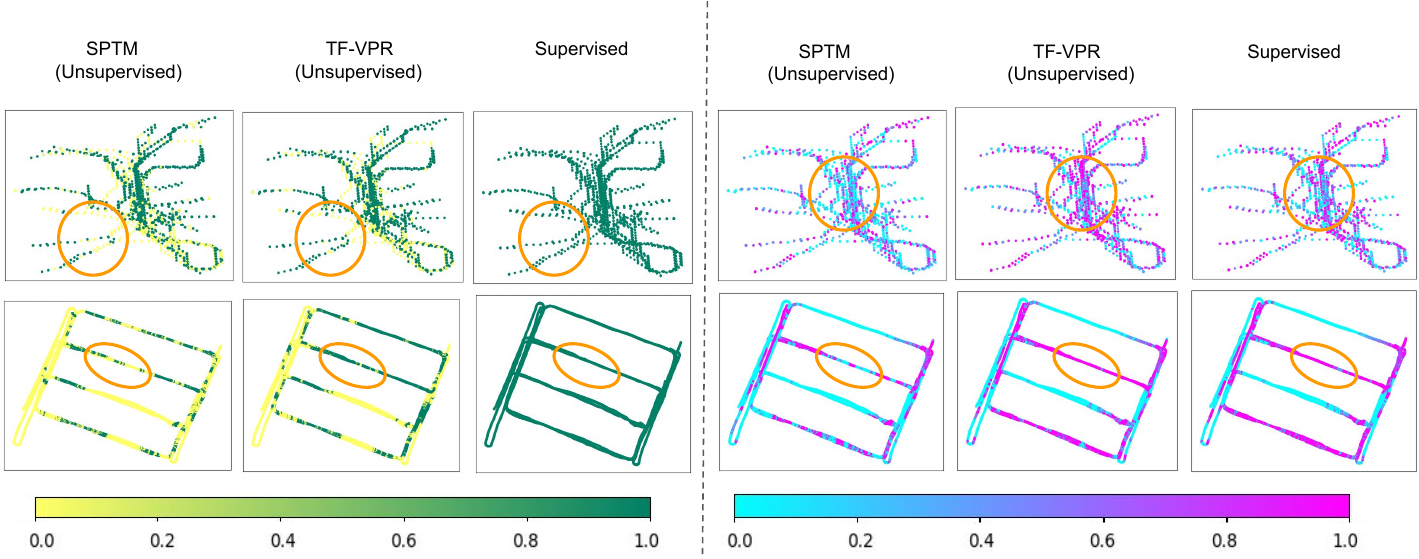}
    \vspace{-5mm}
    \caption{\textbf{Per-frame performance} over ground truth trajectories on point cloud (top), simulated RGB (middle), and NYU-VPR-360 (bottom) datasets. The left block shows the Recall@10 and the right block shows the HD metrics. Each query's metrics at epoch 30 are color-coded.}
    \label{fig:qual_three_datasets}
    \end{center}
\vspace{-5mm}
\end{figure}

\textbf{Comparison with baselines.} TF-VPR's advantage is more evident on the NYU-VPR-360 dataset, as shown by the qualitative results in Fig.~\ref{fig:query_realrgb}. Moreover, Fig.~\ref{fig:qual_three_datasets} demonstrates improvements in recall rate and heading diversity with \titlevariable. Furthermore, as shown in Table~\ref{Table_real}, \titlevariable~surpasses other baselines in recall rate and heading diversity in Scene 1 by about $2\%$ and $4\%$ respectively. Furthermore, performance in Scene 2 does not improve because there are no spatial positives from different headings in Scene 2. More importantly, the performance gap between \titlevariable~and other baselines becomes larger over epochs. In Fig.~\ref{fig:realrgb_exp}, the recall@10 of \titlevariable~outperforms the baselines by approximately \textbf{$8\%$-$10\%$} at epoch $30$ as discussed in Sec.~\ref{CVM}. 

\textbf{Dataset limitation.} While the NYU-VPR-360 dataset includes multiple visits to the same location from various angles, the movement of vehicles on the streets is not completely unrestricted. This limitation affects the effectiveness of augmentation techniques. Additionally, the dataset's overall size is relatively small, which may impact the robustness of our findings. As a result, we present these results in the appendix for reference.

\begin{table}[]
\centering
\captionsetup{font=sc}
\captionsetup{font={scriptsize, sc, stretch=1.3}, justification=centering, labelsep=newline}
\caption{\textbf{Auto-labeling on NYU-VPR-360 dataset.} We employ the best recall rate (R) and heading diversity (HD). Two scenes have been reported (Scene1, Scene2). The best results are highlighted in bold.}
\label{Table_real}
\vspace{-1mm}
\resizebox{1\linewidth}{!}{
\begin{tabular}{cl|cccc|cccc}
\hline
\multicolumn{2}{c|}{Scene}                                              & \multicolumn{4}{c|}{Scene 1}                          & \multicolumn{4}{c}{Scene 2}                          \\ \hline
\multicolumn{2}{c|}{Metric}                                             & \multicolumn{1}{c|}{HD}    & R@1   & R@5    & R@10   & \multicolumn{1}{c|}{HD}    & R@1   & R@5    & R@10   \\ \hline
\multicolumn{2}{c|}{SPTM~\cite{savinov2018semi}}                          & \multicolumn{1}{c|}{79.49} & 48.27 & 60.08  & 69.02  & \multicolumn{1}{c|}{50.79} & \textbf{62.52}  & \textbf{53.25} & 65.97  \\
\multicolumn{2}{c|}{SPTM (epoch 30)~\cite{savinov2018semi}}                          & \multicolumn{1}{c|}{75.46} & 39.34 & 50.27  & 60.16  & \multicolumn{1}{c|}{46.39} & 54.06  & 43.92 & 58.12  \\
\multicolumn{2}{c|}{PCL~\cite{li2020prototypical}}                        & \multicolumn{1}{c|}{0.04}  & 40.99 & 57.30  & 64.75  & \multicolumn{1}{c|}{13.46} & 46.96  & 58.42 & 62.80  \\
\multicolumn{2}{c|}{VLAD~\cite{arandjelovic2013all,jegou2010aggregating}} & \multicolumn{1}{c|}{0.01}  & 10.53 & 27.89  & 37.31  & \multicolumn{1}{c|}{0.01}  & 8.78   & 16.92 & 21.23  \\
\multicolumn{2}{c|}{\titlevariable~(Ours)}                 & \multicolumn{1}{c|}{\textbf{83.34}} & \textbf{52.56}  & \textbf{63.89}  & \textbf{71.94}  & \multicolumn{1}{c|}{\textbf{51.19}} & 62.05  & 52.49 & \textbf{66.16}  \\
\multicolumn{2}{c|}{\titlevariable~(Ours) (epoch 30)}                 & \multicolumn{1}{c|}{80.79} & 50.99 & 62.61  & 70.97  & \multicolumn{1}{c|}{49.33} & 60.09  & 51.44 & 64.18  \\\hline
\multicolumn{2}{c|}{NetVLAD~\cite{arandjelovic2016netvlad}}               & \multicolumn{1}{c|}{86.38} & 100.00 & 100.00 & 100.00 & \multicolumn{1}{c|}{54.45} & 99.99  & 100.00 & 100.00 \\ \hline
\end{tabular}
}
\vspace{-6mm}
\end{table}

\vspace{-2mm}

 \vspace{-6mm}

\begin{figure*}[ht]
    \centering {\includegraphics[trim={0.7cm 15cm 0.5cm 0.5cm},clip,width=0.96\textwidth
    ]{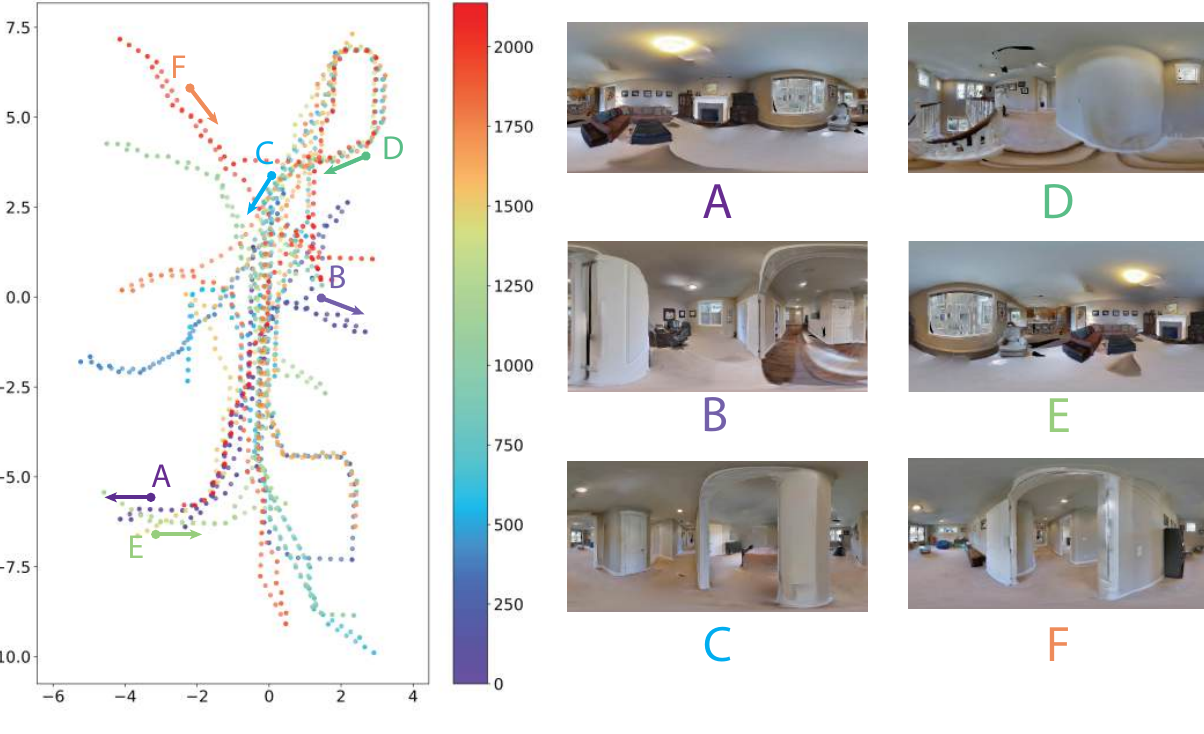}} \quad
    \vspace{-4mm}
    \caption{\textbf{Qualitative demonstrations of Habitat-Sim dataset}. Each point on the trajectory is assigned a color to represent its timestamp. The arrows at each point indicate their orientations. The images captured at several trajectory points are visualized.}
    \label{fig:goffs}
    
\end{figure*}

\begin{figure*}[t]
    \centering {\includegraphics[trim={0.7cm 13cm 0.5cm 0.5cm},clip,width=0.96\textwidth
    ]{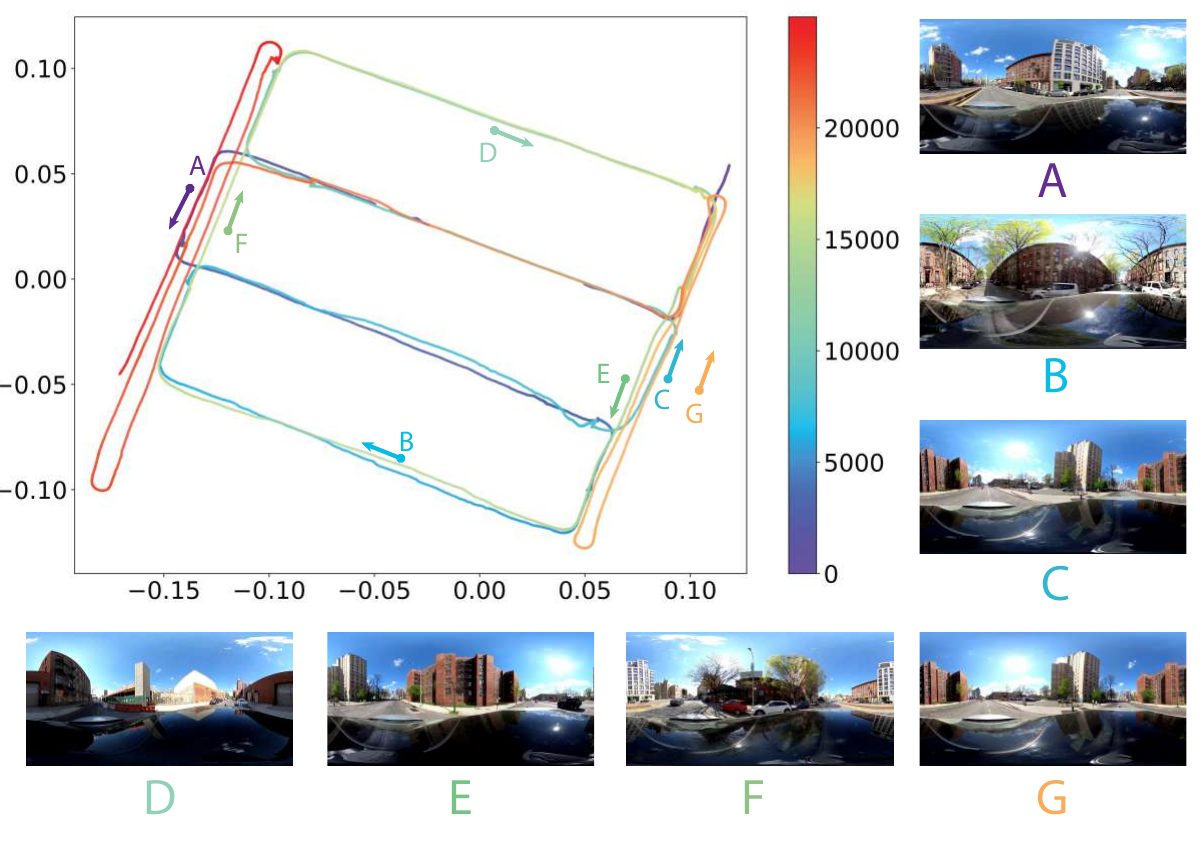}} \quad
    \vspace{-5mm}
    \caption{\textbf{Qualitative demonstrations of NYU-VPR-360 dataset}. The legend follows Fig.{~\ref{fig:goffs}}}
    \label{fig:real-rgb}
\end{figure*}

\vspace{7mm}
\section{Dataset}
\textbf{Habitat-Sim dataset.} We collected the photo-realistic simulated images in the Habitat-Sim~\cite{savva2019habitat} simulator using the Gibson~\cite{xiazamirhe2018gibsonenv} dataset. The RGB images were captured by a virtual 360 camera mounted on a virtual robot in the environment. The robot moved according to the random exploration strategy. In total, we collected more than 10k images in 18 scenes. Fig~\ref{fig:goffs} shows some examples of images and robot trajectories.

\clearpage